\documentclass[10pt,letterpaper]{article}

\usepackage[margin=1in]{geometry}
\usepackage{times}
\usepackage{microtype}
\usepackage{enumitem}
\setlist[itemize]{leftmargin=*, itemsep=0pt, topsep=2pt}

\usepackage{graphicx}
\usepackage{xcolor}
\usepackage{amsmath,amssymb}
\usepackage{booktabs}
\usepackage{caption}
\usepackage{setspace}

\providecommand{\medium}{}

\usepackage{graphicx}
\usepackage{subcaption}
\usepackage{adjustbox}
\usepackage{float}
 \usepackage{multirow}

 \usepackage{setspace}
\usepackage{placeins}

\usepackage{xcolor}
\usepackage{tikz}
\usetikzlibrary{plotmarks,positioning,calc,shapes}
\usepackage{pgfplots}
\pgfplotsset{compat=1.18}

\usepackage{etoolbox}
\AtBeginEnvironment{thebibliography}{\medium}

\usepackage{pifont}
\definecolor{darkgreen}{RGB}{0,100,0}
\newcommand{\cmark}{\textcolor{darkgreen}{\ding{51}}} 
\newcommand{\xmark}{\textcolor{red}{\ding{55}}} 
\usepackage{adjustbox}
\usepackage[table]{xcolor}

\usepackage[square,numbers,sort&compress]{natbib}
\usepackage[hidelinks]{hyperref}

\title{HERBench: A Benchmark for Multi-Evidence Integration in Video Question Answering}

\author{%
\begin{tabular}{c}
\textbf{Dan Ben-Ami}$^{1,*}$ \quad
\textbf{Gabriele Serussi}$^{1,*}$ \quad
\textbf{Kobi Cohen}$^{2}$ \quad
\textbf{Chaim Baskin}$^{1}$\\[0.35em]
{\small $^{1}$INSIGHT Lab, Ben-Gurion University of the Negev, Israel}\\
{\small $^{2}$Ben-Gurion University of the Negev, Israel}\\[0.25em]
{\small $^{*}$Equal contribution.}\\
\end{tabular}%
}

\date{} 

\begin{document}
\maketitle

\begin{figure}[t]
    \centering
    \includegraphics[width=\textwidth]{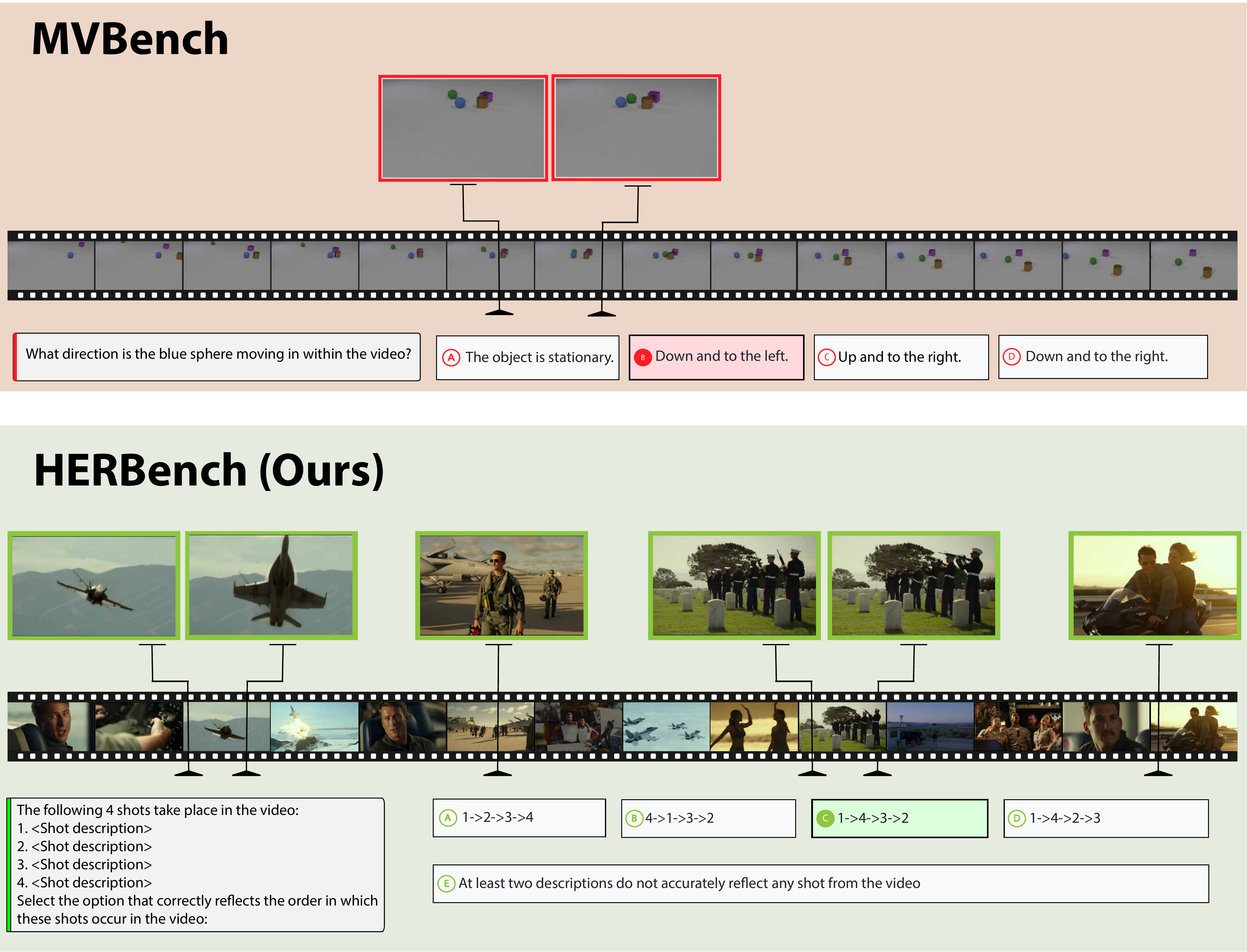}
    \caption{\textbf{From Single-Cue to Multi-Evidence Integration.} While existing benchmarks like MVBench \cite{li2024mvbench} (top) often focus on short-term attributes solvable via single salient frames or language priors, \textbf{HERBench} (bottom) enforces a high Evidential Requirement (ER). In this \textit{Temporal Shot Ordering} example, the model must identify and temporally bind four distinct, non-overlapping visual evidence dispersed across the video to reconstruct the correct sequence. This design ensures that successful answering requires genuine multi-evidence integration rather than reliance on static shortcuts.}
    \label{fig:teaser}
\end{figure}

\begin{abstract}
Video Large Language Models (Video-LLMs) are improving rapidly, yet current Video Question Answering (VideoQA) benchmarks often admit single-cue shortcuts, under-testing reasoning that must integrate evidence across time. We introduce HERBench, a benchmark designed to make multi-evidence integration unavoidable: each question requires at least three non-overlapping cues drawn from distinct video segments. HERBench contains 26,806 five-way multiple-choice questions across 12 compositional tasks. To make evidential demand measurable, we introduce the Minimum Required Frame-Set (MRFS), the smallest number of frames a model must fuse to answer correctly, and show that HERBench imposes higher evidential demand than prior benchmarks. Evaluating 13 state-of-the-art Video-LLMs yields only 31--42\% accuracy, only modestly above the 20\% random-guess baseline. We disentangle this failure into two critical bottlenecks: (1) a retrieval deficit, where frame selectors overlook key evidence, and (2) a fusion deficit, where models fail to integrate information even when all necessary evidence is provided.
HERBench thus provides a principled benchmark for studying robust multi-evidence video understanding. \href{https://herbench.github.io/}{\includegraphics[height=1em]{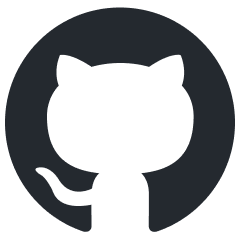}}
\end{abstract}
\section{Introduction}
As Video Large Language Models~\cite{bai2025qwen3vl, llava_onevision15_8b_2025, ovis25_9b_2025} achieve strong scores on established VideoQA benchmarks~\cite{lei2018tvqa, jang2017tgif, xiao2021nextqa, tapaswi2016movieqa, mangalam2023egoschema, fu2024videomme, longvideobench2024}, their video understanding capabilities appear to be rapidly emerging.
However, recent audits reveal these high scores often stem from language priors or single-cue shortcuts rather than grounded temporal reasoning~\cite{tempcompass2024, xiao2024visground}, causing models to fail tasks that explicitly require multi-hop inference~\cite{girdhar2020cater, yi2020clevrer, grundemclaughlin2021agqa}.
In contrast, tasks like Referring Video Object Segmentation (RVOS) demonstrate that robust, multi-frame aggregation is achievable, as models successfully link instances across occlusions and appearance changes~\cite{Gavrilyuk_2018_CVPR, Seo_2020_ECCV, Botach_2022_CVPR, Wu_2022_CVPR}.

We advocate centering evaluation on \textit{evidential requirement}, because single-cue questions fail to measure multi-evidence integration. We define the \textbf{Evidential Requirement (ER)} as the minimum number of distinct, non-redundant visual evidence needed for an answer. High-ER items make compositional reasoning, such as temporal binding and clue combination, unavoidable~\cite{xiao2021nextqa, xiao2021star, yi2020clevrer}. Controlling ER therefore distinguishes models that integrate information from those that rely on isolated cues~\cite{tempcompass2024}. This approach makes aggregation measurable, aligns VideoQA with real-world reasoning, and offers a principled path for progress beyond single-cue success~\cite{xiao2021nextqa, xiao2021star, yi2020clevrer}.

We introduce \textbf{HERBench} (High Evidential Requirement Benchmark), where questions across twelve compositional subtasks (e.g., entity binding, temporal ordering) are constructed to \textit{ structurally enforce} $k \geq 3$ distinct pieces of evidence, as presented in Figure~\ref{fig:teaser}. To measure this, we present the \textbf{Minimum Required Frame-Set (MRFS)} metric, defined as the minimum number of frames needed for a correct answer. Cross-benchmark comparison under a canonical MRFS protocol confirms our high-ER design: HERBench attains the highest mean MRFS among the benchmarks considered and enables principled ER-focused diagnostics.

Our evaluation of state-of-the-art Video-LLMs exposes \textit{two critical bottlenecks}.
\textbf{Finding 1:} Frame selection is a major bottleneck. While adaptive selectors~\cite{tang2025aks, liu2025bolt} outperform uniform sampling, they still lag significantly behind ground-truth keyframes.
\textbf{Finding 2:} Multi-evidence reasoning is also a bottleneck. Even with ground-truth frames, models achieve only modest accuracy because they \textit{fail to assign proper importance to all critical frames} and struggle to integrate them. Progress therefore requires advances in both frame selection and multi-evidence reasoning.

\noindent\textbf{Our main contributions are summarized as follows.}
\begin{itemize}[leftmargin=*, itemsep=0pt, topsep=2pt]
\item \textbf{We introduce \emph{HERBench}:} a benchmark with 26{,}806 questions that are \emph{constructed to structurally enforce} $k \geq 3$ distinct, non-redundant visual cues.
\item \textbf{We propose the \emph{Minimum Required Frame-Set} (MRFS) metric:} a measure of the smallest number of frames a model must aggregate to answer a question correctly, thereby enabling apples-to-apples comparison across benchmarks and powering ER-focused diagnostics.
\item \textbf{We identify two critical bottlenecks in current Video-LLMs:} By disentangling \emph{frame selection} from \emph{multi-evidence reasoning}, we reveal two systemic failures. (i) \emph{Frame selection:} Adaptive selectors, though an improvement over uniform sampling, still overlook key evidence and do not yet match the performance of oracle key-frames. (ii) \emph{Multi-evidence reasoning:} Even with oracle frames, models fail to integrate complementary information and systematically underweight necessary evidence. Progress requires advances in both selection and reasoning.
\end{itemize}
\section{Related Work}
\paragraph{Video Large Language Models.}
Video-LLMs architectures have evolved from simple feature pooling \cite{damonlpsg2023videollama,Maaz2023VideoChatGPT} to sophisticated systems employing advanced alignment modules (e.g., Q-Formers) and large-scale instruction tuning \cite{li2023blip2, dai2023instructblip, llava_onevision_7b_2024} to bridge the modality gap. Despite massive token capacities in proprietary models like Gemini 2.5 \cite{comanici2025gemini} and GPT-4o \cite{openai2024gpt4o}, recent audits \cite{tempcompass2024, breaking-down} reveal a persistent failure in robust temporal aggregation. Instead of performing multi-hop inference, these models frequently default to language priors or single-frame shortcuts to solve tasks.

\paragraph{Video Question Answering Benchmarks.}
VideoQA benchmarks have evolved from short-clip recognition~\cite{xu2017video,xu2016msrvtt} toward longer-form evaluation, but they often probe different aspects of video understanding. MVBench~\cite{li2024mvbench} broadened the task space with diverse temporal questions, yet its short clips limit the assessment of long-horizon assessment. More recent benchmarks, including EgoSchema~\cite{mangalam2023egoschema}, LongVideoBench~\cite{longvideobench2024}, and Video-MME~\cite{fu2024videomme}, expand temporal scope, while MINERVA~\cite{nagrani2025minerva} emphasizes complex multi-step reasoning and reasoning-trace auditing. HERBench targets a different axis: not only how long the context is, but how many visual pieces of evidence must be combined. While phenomena such as temporal ordering and counting appear in prior benchmarks, they are typically not formulated so that solving them \emph{requires} aggregating multiple temporally separated cues. HERBench instead explicitly controls the \textit{Evidential Requirement (ER)}: each question is constructed to require at least three non-overlapping cues drawn from distinct moments in the video, and the benchmark's oracle-frame design enables retrieval failures to be disentangled from fusion failures.
\begin{figure*}[t]
  \centering
  \includegraphics[width=\textwidth]{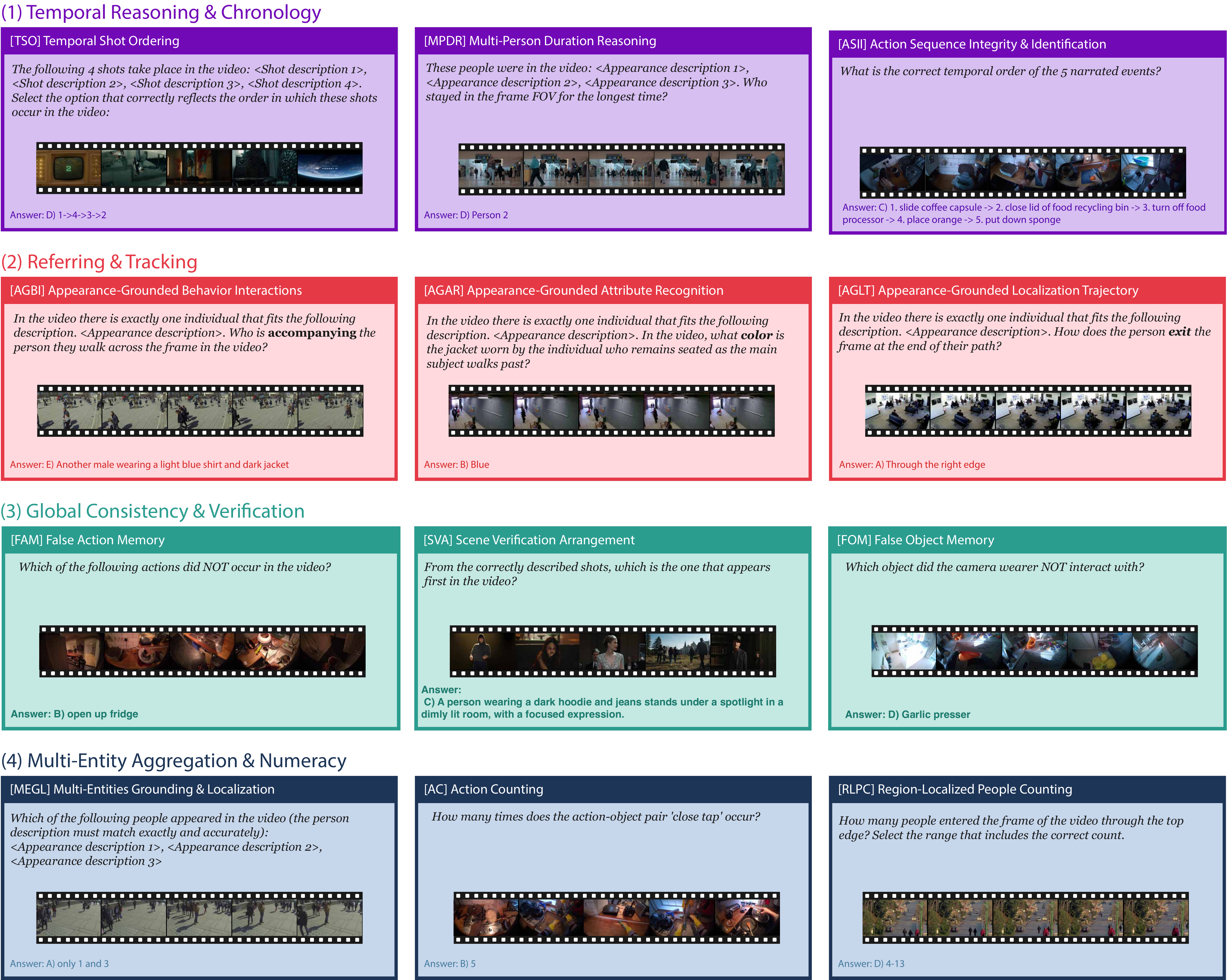}
  \caption{\textbf{Task taxonomy of HERBench.} We organize 12 fine-grained compositional tasks into four essential reasoning families: \textbf{(1) Temporal Reasoning \& Chronology}, \textbf{(2) Referring \& Tracking}, \textbf{(3) Global Consistency \& Verification}, and \textbf{(4) Multi-Entity Aggregation \& Numeracy}. Unlike existing benchmarks that may allow for single-frame shortcuts, every task in HERBench is constructed to enforce a \textbf{High Evidential Requirement}, requiring models to aggregate at least three distinct, temporally separated visual cues ($k \ge 3$) to derive the correct answer.}
  \label{fig:grid_plot}
\end{figure*}

\section{HERBench: High Evidential Requirement Benchmark}
\label{sec:method}

\subsection{Task Taxonomy}
\label{subsec:tasks}
To evaluate whether models truly integrate evidence rather than rely on a salient cue, we organize 12 tasks into four reasoning families (Figure~\ref{fig:grid_plot}). These families recast familiar VideoQA settings under structural $k \geq 3$ constraints over long videos, while also introducing tasks for appearance-grounded identity binding, set-level identity maintenance, and region-conditioned aggregation.

\paragraph{Temporal Reasoning \& Chronology [TR\&C].}
These tasks require understanding event order, co-occurrence, and durations, compiling distributed cues into a linear chronology. The three tasks are: 1) \textit{[TSO] Temporal Shot Ordering}: Arrange four shot descriptions from a trailer into the correct chronological order, using only content cues. 2) \textit{[MPDR] Multi-Person Duration Reasoning}: Compare interval statistics for appearance-described people (e.g., who stayed in view the longest, or who entered/exited first), focusing on \emph{fine-grained time-span contrasts} across individuals. 3) \textit{[ASII] Action Sequence Integrity \& Identification}: Select the correct ordering of five narrated actions among plausible permutations, stressing \emph{micro-level task sequencing} rather than scene-level ordering. The ER is driven by ordering and interval comparisons across at least three temporally separated observations, but each task probes a distinct temporal structure.

\paragraph{Referring \& Tracking [R\&T].}
This family tests binding a uniquely appearance-described target across time to reason about trajectory-dependent properties. Models must maintain a stable reference as the target interacts with the scene. The tasks are: 1) \textit{[AGBI] Appearance-Grounded Behavior Interactions}: Identify who accompanies or interacts with the target during traversal, emphasizing \emph{social and relational} cues. 2) \textit{[AGAR] Appearance-Grounded Attribute Recognition}: Track the target to read out attributes anchored to their immediate local context (e.g., a passerby’s jacket color), focusing on \emph{moment-specific attribute extraction}. 3) \textit{[AGLT] Appearance-Grounded Localization Trajectory}: Recover path endpoints and coarse trajectory (e.g., exit method), highlighting \emph{global, path-level motion reasoning}. This enforces $k \ge 3$ through identity maintenance across separated glimpses,  as the target description is composed of cues scattered across distinct moments that must be jointly resolved, with each task centering on a different aspect of target evolution.

\paragraph{Global Consistency \& Verification [GC\&V].}
Next, we test exhaustive video-wide verification and absence detection, sweeps that must confirm what occurred and surface plausible but missing elements. The three tasks are: 1) \textit{[FAM] False Action Memory}: Among several plausible actions, select the one that never occurs while verifying the others do, requiring \emph{action-level absence detection}. 2) \textit{[SVA] Scene Verification Arrangement}: Given 2-4 shot descriptions where some may be fabricated, first identify the faithful ones, then arrange the correct shots in temporal order, or return a calibrated abstention when too many descriptions are false; this combines \emph{shot-level fidelity checking} with chronology. 3) \textit{[FOM] False Object Memory}: Among plausible objects, identify the one the camera wearer does \emph{not} interact with while verifying the rest, stressing \emph{object-level absence} tied to first-person interactions. Here $k \ge 3$ arises from multi-moment sweeps needed to validate presence and detect absence across the video.

\paragraph{Multi-Entity Aggregation \& Numeracy [MEA\&N].}
Finally, this family stresses many-way binding, spatial partitioning, and precise counting across multiple people or events. Models must deduplicate identities across time and fuse evidence spread over the video. The three tasks are: 1) \textit{[MEGL] Multi-Entities Grounding \& Localization}: Given 2-3 detailed appearance descriptions, decide which individuals actually appear in the video (exact-match verification among plausible distractors), focusing on \emph{set membership and identity deduplication}. 2) \textit{[AC] Action Counting}: Count the occurrences of a specified action-object pair distributed across the timeline, emphasizing \emph{event-accumulation across dispersed moments}. 3) \textit{[RLPC] Region-Localized People Counting}: Count unique individuals subject to spatial constraints (e.g., entries through the top edge), with answers reported as binned ranges, requiring \emph{region-conditioned identity aggregation}. Here $k \ge 3$ is enforced by set-level aggregation and cardinality constraints over multiple moments, with each task stressing a complementary aggregation mode.

\subsection{Benchmark Construction}
\label{subsec:benchmark_construction}
\begin{figure*}[t!]
  \centering
  \includegraphics[width=\textwidth]{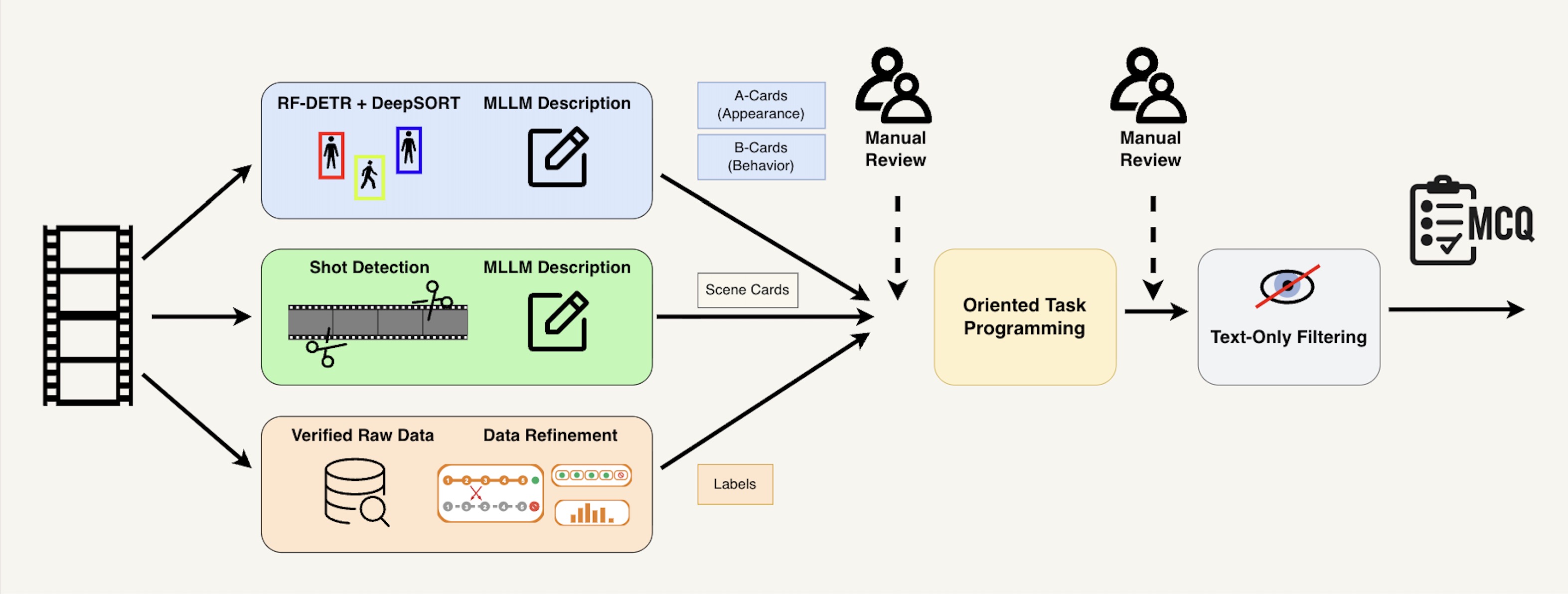}
  \caption{\textbf{HERBench construction pipeline.} Videos are processed through three streams---object tracking and trajectory analysis, shot segmentation, and ground-truth integration (refining human verified raw event logs) ---whose outputs are compiled by oriented task programming and filtered by manual review and text-only bias checks.}
  \label{fig:data_pipeline}
\end{figure*}

We construct HERBench through the tripartite data construction pipeline shown in Figure~\ref{fig:data_pipeline}. The core of this process is the creation of a rich spatiotemporal scaffold by processing each video through three complementary streams. The first stream, \textit{Object Tracking \& Trajectory Analysis}, focuses on continuous, micro-level object dynamics. Complementing this, the second stream, \textit{Shot Segmentation}, provides a macroscopic view by discretizing the video into semantic units. Finally, the \textit{Ground Truth Integration} stream anchors the analysis in human-verified facts. Together, these streams produce a diverse set of refined data (such as A/B cards, scene cards, and event labels).

\paragraph{Pipeline I: Object Tracking \& Trajectory Analysis.}
This first stream anchors tasks in continuous object dynamics. We employ RF-DETR \cite{robinson2025rfdetr} and DeepSORT \cite{wojke2017simple} to obtain entity tracks, retaining top 20\% entities via a \textit{TrackRank} score, a composite score favoring appearance rarity, trajectory length and frame coverage (see Supplementary). For each track, we generate strictly non-overlapping \textit{A-cards} (appearance) and \textit{B-cards} (behavior/trajectory). This decorrelation intentionally separates the identifying appearance from the queried behavior, often placing them in temporally distant frames, and enables a new class of appearance-grounded identity-binding tasks that cannot be solved by local attribute lookup alone. Full error quantification and noise control details are provided in the Supplementary.

This scaffold supports tasks requiring fine-grained interaction and motion analysis:
\begin{itemize}[leftmargin=*]
    \item \textbf{[AGBI], [AGAR], [AGLT]:} We generate questions strictly from \textit{B-cards} while referring to entities via \textit{A-cards}, separating appearance from behavior. \textit{[AGBI]} queries behavioral interactions with other entities; \textit{[AGAR]} queries attributes; and \textit{[AGLT]} queries path integration and motion topology.
    \item \textbf{[MPDR]:} We compute per-entity visible-time intervals to generate queries comparing durations (e.g., longest presence) or checking for temporal overlaps. The correct answer is, by definition, a property of the relationship between multiple, ordered cues.
    \item \textbf{[RLPC]:} We execute spatial programs to count unique track IDs traversing predefined regions of interest or entry/exit gates, testing spatiotemporal aggregation capabilities.
    \item \textbf{[MEGL]:} We form sets of appearance descriptors and inject plausible distractors, forcing models to verify the exact set of present individuals throughout the video.
\end{itemize}

\paragraph{Pipeline II: Shot Segmentation.}
Where the first pipeline focuses on continuous entity-level detail, this second stream discretizes the video into larger semantic units. It uses shot boundary detection, employing an MLLM to summarize each segment into a concise scene card. 
This macroscopic view supports tasks dependent on global temporal coherence:
\begin{itemize}[leftmargin=*]
    \item \textbf{[TSO]:} We query the chronological arrangement of the generated scene cards, requiring the model to reorder shuffled scenes.
    \item \textbf{[SVA]:} We mix faithful scene cards with plausibly perturbed variants, altering 2-5 atomic details (e.g. actions, attributes), to test resistance to gist cues or partially correct descriptions.
\end{itemize}

\paragraph{Pipeline III: Ground Truth Integration.}
Finally, this stream moves beyond automated analysis to leverage human verified narrated events \cite{perrett2025hdepic}:
\begin{itemize}[leftmargin=*]
    \item \textbf{[FAM], [FOM]:} We introduce corpus-plausible distractors, entities or actions common in similar videos but verified as absent, requiring multi-timestamp scanning rather than single-frame spot checks.
    \item \textbf{[ASII]:} We establish ground-truth chronology from narrated events and present proposed sequences (faithful vs. perturbed) for careful verification.
    \item \textbf{[AC]:} Ground-truth counts are derived directly from verified event logs to test long-horizon aggregation.
\end{itemize}


\paragraph{Synthesis \& Quality Control.} The refined representations are compiled by oriented task programming into multiple-choice questions, then filtered through several safeguards. We enforce A/B-card disentanglement via token-level similarity checks and manual leakage review, discard items solved by $\geq 3$ of 4 blind LLMs, and perform expert verification on a stratified 15\% sample to confirm minimum frame-set ($k\ge3$) compliance and answer uniqueness. Human validation further confirmed both answerability and oracle-evidence quality: annotators achieved 88.8\% accuracy with full-video access and 95.7\% in an oracle-frame setting; details are deferred to the Supplementary Material. As HERBench is partially instantiated through an oriented task-programming pipeline, some residual systematic artifacts may nevertheless remain.

\newcommand{\wordcloudscale}{0.32}
\newcommand{\piechartscale}{0.25}
\newcommand{\histogramscale}{0.41}

\begin{figure*}[t]
    \centering
    \begin{subfigure}{\wordcloudscale\textwidth}
        \centering
        \includegraphics[width=\linewidth, keepaspectratio]{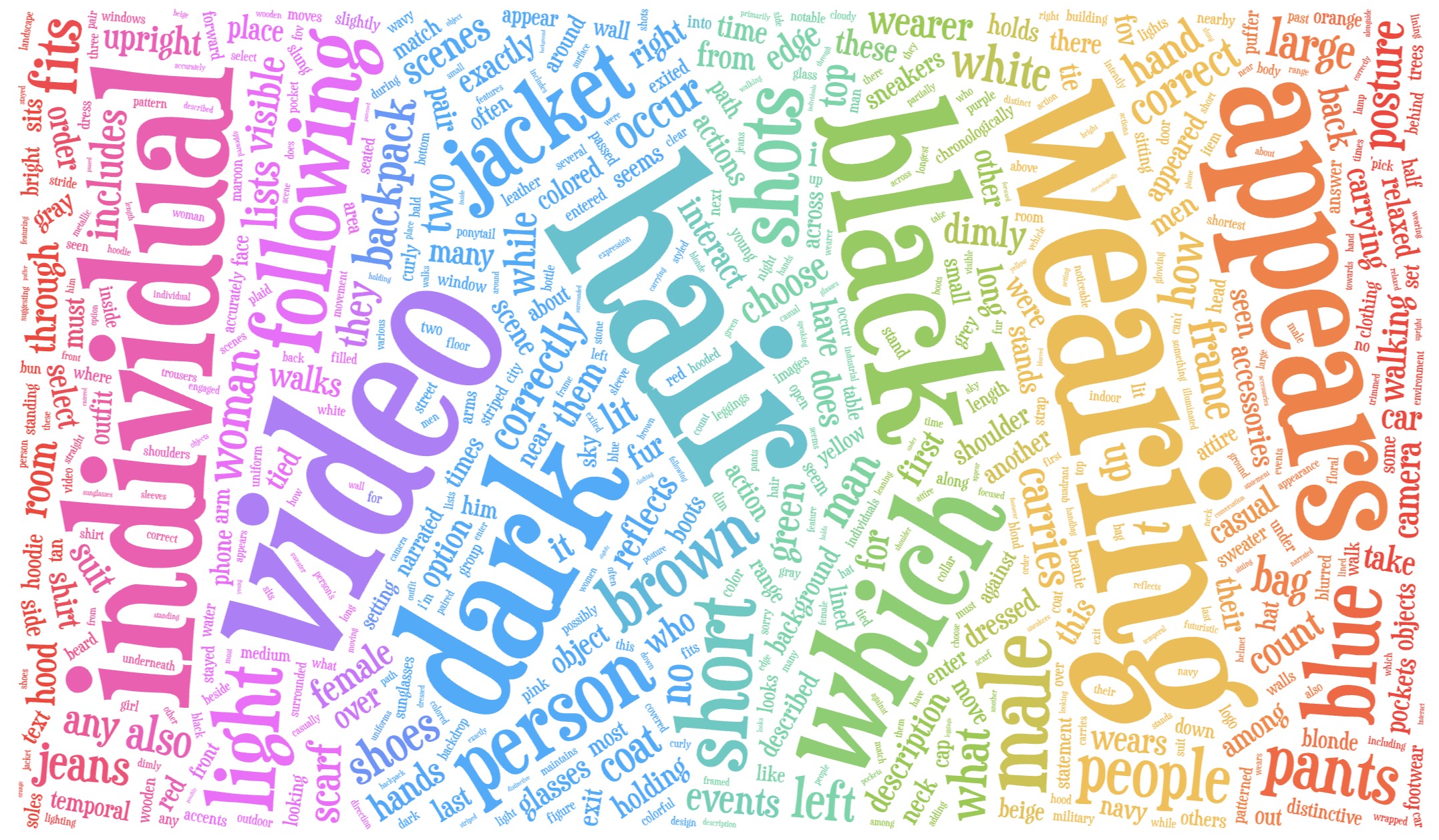}
    \end{subfigure}
    \hfill
    \begin{subfigure}{\piechartscale\textwidth}
        \centering
        \includegraphics[width=\linewidth, keepaspectratio]{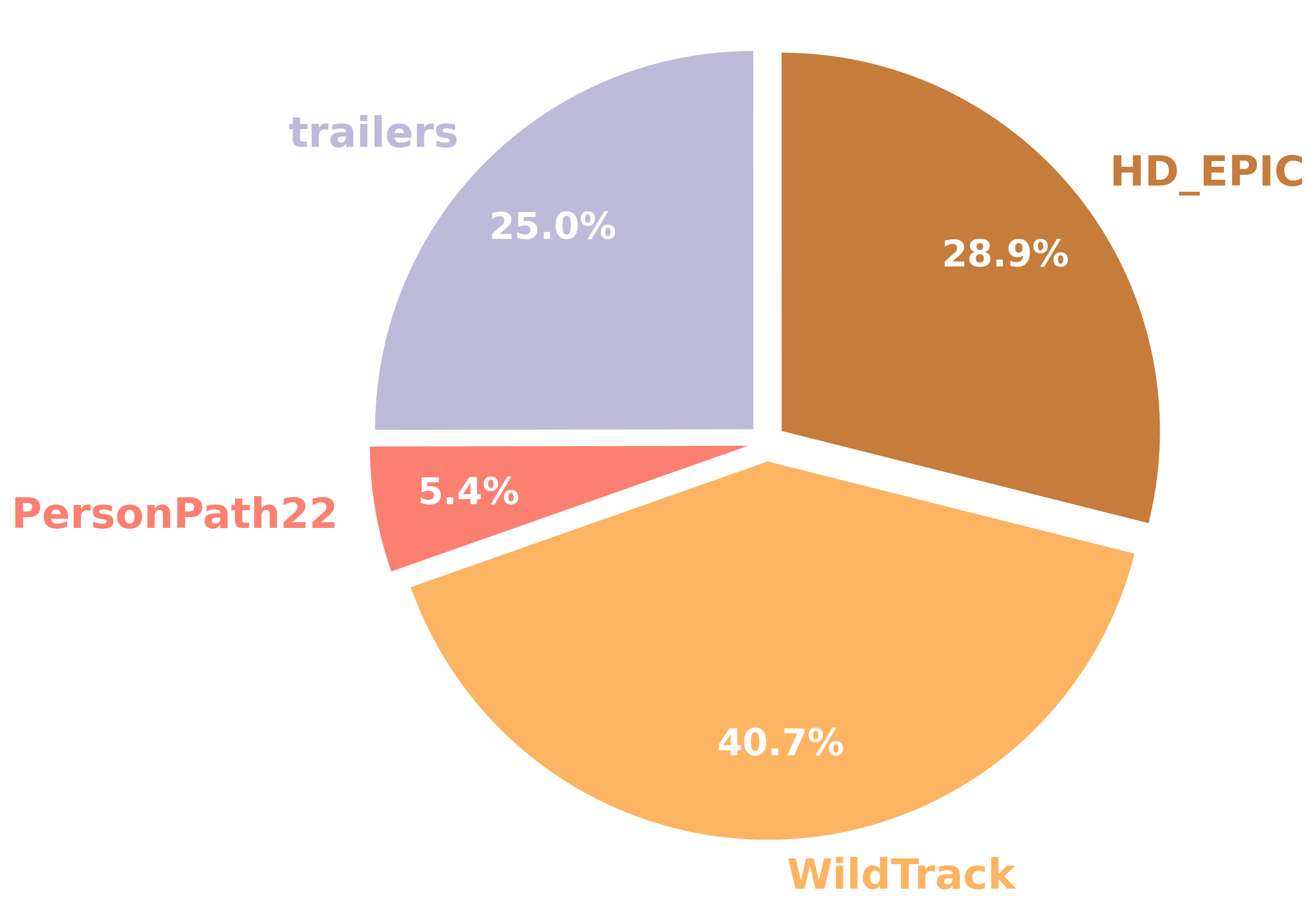}
    \end{subfigure}
    \hfill
    \begin{subfigure}{\histogramscale\textwidth}
        \centering
        \includegraphics[width=\linewidth, keepaspectratio]{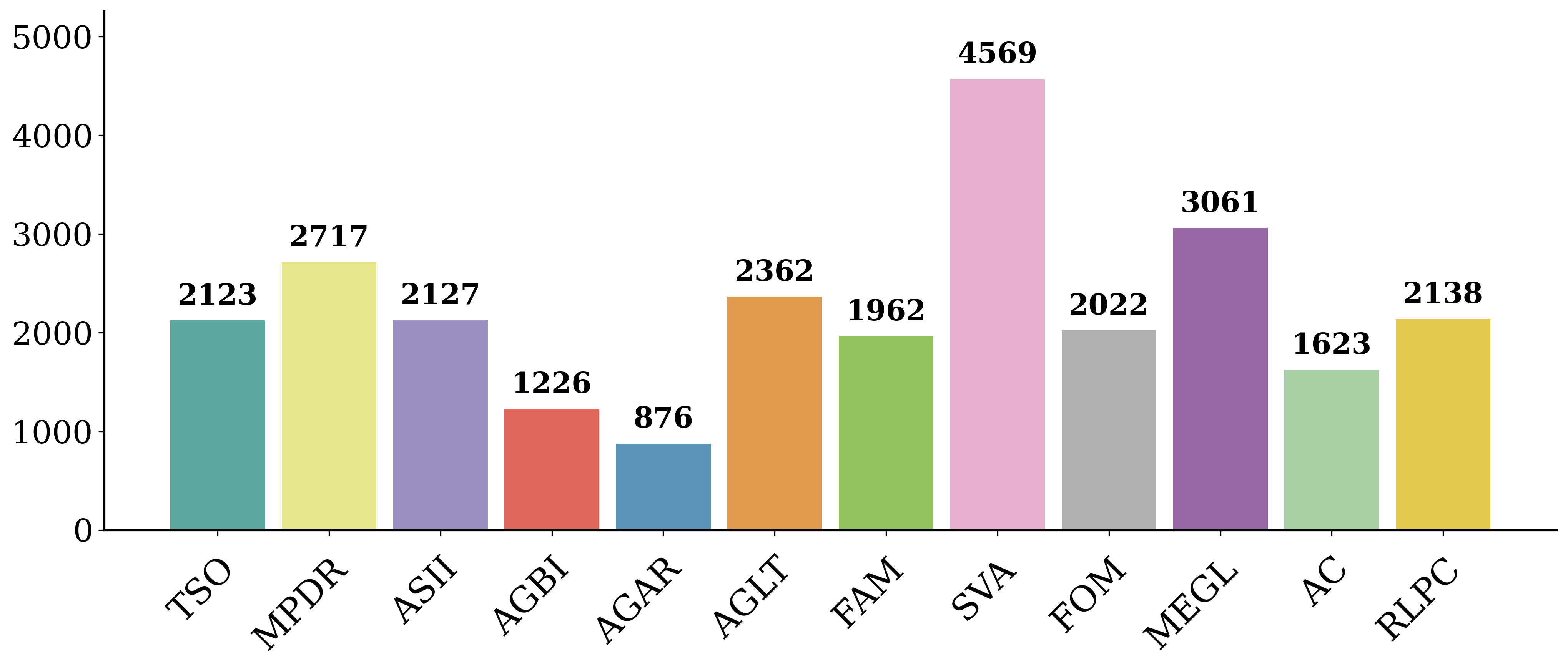}
    \end{subfigure}

    \caption{ \textbf{Left:} Wordcloud of frequent terms in HERBench queries. \textbf{Center:} Distribution of samples across source datasets. \textbf{Right:} Number of questions per task category.}
    \label{fig:word_pie_hist}
\end{figure*}

\subsection{Benchmark Statistics}
\label{subsec:stats}




HERBench contains 26,806 five-way multiple-choice questions from 336 unique videos spanning 12 tasks and four reasoning families. The videos are long-form (395\,s on average; range 60--2100\,s), diverse in source and viewpoint, and chosen to ensure temporal dispersion of evidence; Figure~\ref{fig:word_pie_hist} summarizes the resulting vocabulary, source distribution, and per-task question counts.

\subsection{Evidential Requirement \& the MRFS Metric}
\label{subsec:mrfs}

\paragraph{Motivation.}
VideoQA questions may require fusing distributed evidence or be solvable via single salient frames. To quantify this, we introduce the \textbf{Minimum Required Frame-Set (MRFS)}: the smallest number of frames a model must fuse to answer correctly. A higher mean MRFS confirms that questions resist single-cue shortcuts and genuinely demand multi-moment integration. Although prior metrics capture important temporal dynamics, they measure fundamentally different properties. Temporal Indispensability (1-frame vs. N-frame performance) \cite{mmbenchvideo2024} tests for static shortcuts, while Certificate Length \cite{fu2024videomme, mangalam2023egoschema} measures the human-annotated temporal span needed for verification. In contrast, MRFS is an automated, model-centric metric that isolates the exact number of frames a model must fuse, directly quantifying the multi-evidence aggregation challenge without relying on human annotation.

\paragraph{Definition.}
Let $v$ denote the video, $q$ the question, and $y$ the ground-truth answer. Let $f$ be a fixed MLLM, $r$ a question-conditioned frame selector, and $x$ a frame budget. The selector produces a ranking $\pi=r(v,q)$ over frames and we denote $F_k=\{\pi_1,\ldots,\pi_k\}$ as the top-$k$ subset. With evaluator $E(\hat{y},y)=\mathbf{1}\{\hat{y}=y\}$, we define $\mathrm{MRFS}_x(q; f, r)\;=$
\begin{equation}
\;\min\bigl\{\,k \in \{1,\dots,x\}\;:\; E\!\bigl(f(q, F_k),\,y\bigr)=1\,\bigr\},
\label{eq:mrfs}
\end{equation}
subject to the precondition $E(f(q,\varnothing),y)=0$ so that text-only solvable items are excluded from MRFS computation. Intuitively, $\mathrm{MRFS}_x$ is the \emph{least} amount of visual evidence (in frames) that suffices for $f$ to be correct when frames are supplied in an $r$-determined, question-aware order.

\paragraph{Computation.}
We search for the smallest success index using an \emph{adaptive bisection} over $k\in[1,x]$, requiring $O(\log x)$ model calls per item. Each question is categorized as: (i) \emph{text-only} (correct with no frames, $f(q, \varnothing)$), (ii) \emph{visual-required} (correct for some $1 \le k \le x$), or (iii) \emph{undefined} (incorrect even at $k=x$).

\begin{table}[t]
\centering
\vspace{-0.25cm}
\small
\setlength{\tabcolsep}{5pt}
\resizebox{\linewidth}{!}{
\begin{tabular}{lccccccc}
\toprule
\textbf{Benchmark} &
\textbf{\begin{tabular}[c]{@{}c@{}}\# Videos\end{tabular}} &
\textbf{\begin{tabular}[c]{@{}c@{}}\# Questions\end{tabular}} &
\textbf{MRFS$\uparrow$} &
\textbf{\begin{tabular}[c]{@{}c@{}}Lang.\\debias\end{tabular}} &
\textbf{\begin{tabular}[c]{@{}c@{}}Enforced\\fusion\end{tabular}} &
\textbf{\begin{tabular}[c]{@{}c@{}}Absence\\check\end{tabular}} &
\textbf{\begin{tabular}[c]{@{}c@{}}Oracle\\frames\end{tabular}} \\
\midrule
TemporalBench   & 2{,}179 & 9{,}867  & 2.21 & \xmark & \xmark & \xmark & \xmark \\
MMBench-Video   & 609     & 1{,}998  & 4.41 & \xmark & \xmark & \xmark & \xmark \\
Video-MME       & 900     & 2{,}700  & 5.31 & \xmark & \xmark & \xmark & \xmark \\
AGQA (balanced)     & 9{,}600 & 3.9 M     & 3.42 & \xmark & \xmark & \xmark & \xmark \\
CVRR-ES         & 217     & 2{,}400  & 2.77 & \xmark & \xmark & \cmark & \xmark \\
LongVideoBench  & 3{,}763 & 6{,}678  & 4.07 & \xmark & \xmark & \xmark & \xmark \\
NExT-QA         & 5{,}440 & 99{,}736 & 2.61 & \xmark & \xmark & \xmark & \xmark \\
MINERVA         & 223     & 1{,}515  & 5.14 & \cmark & \xmark & \xmark & \xmark \\
MVBench         & 4{,}000 & 4{,}000  & 3.52 & \xmark & \xmark & \xmark & \xmark \\
\midrule
\textbf{HERBench} & 336 & 26,806 & \textbf{5.49} & \cmark & \cmark & \cmark & \cmark \\
\bottomrule
\end{tabular}}
\caption{\textbf{Benchmark comparison under the canonical MRFS protocol.} MRFS is reported with $f=$ Qwen2.5-VL, $r=$ AKS, and $x=16$. \emph{Enforced fusion} indicates whether correct answering structurally requires combining multiple temporally separated visual cues.}
\label{tab:benchmark-stats}
\end{table}

\paragraph{Cross-benchmark MRFS comparison.}
To enable cross-benchmark comparison, we fix a canonical MRFS protocol: Qwen2.5-VL as backbone, AKS as frame selector, and $x=16$ frames. Table~\ref{tab:benchmark-stats} compares HERBench to prior benchmarks under this shared setting, together with complementary design properties. HERBench achieves the highest mean MRFS (\textbf{5.49}) in the comparison and is uniquely defined by the combination of language debiasing, structurally required temporal fusion, absence checking, and oracle-frame availability. Among prior benchmarks, MINERVA \cite{nagrani2025minerva} is the closest in spirit, pairing language debiasing with a relatively high MRFS (\textbf{5.14}), but focusing on multi-step reasoning and reasoning-trace auditing rather than enforced multi-frame evidence aggregation. The benchmark ordering remains stable across alternative backbones and selectors (see Supplementary), and the comparison with LongVideoBench \cite{longvideobench2024} suggests that HERBench derives its difficulty from evidential density rather than duration alone.
\section{Experiments}
\label{sec:experiments}
\begin{table*}[t]
\centering
\renewcommand{\arraystretch}{1.25} 
\setlength{\tabcolsep}{3.5pt}      
\begin{adjustbox}{max width=\textwidth}
{\large
\begin{tabular}{l|ccc|c|ccc|c|ccc|c|ccc|c|c}
\toprule
\multirow{2}{*}{\textbf{Model}} &
\multicolumn{4}{c|}{\textbf{TR\&C}} &
\multicolumn{4}{c|}{\textbf{R\&T}} &
\multicolumn{4}{c|}{\textbf{GC\&V}} &
\multicolumn{4}{c|}{\textbf{ME\&N}} &
\multirow{2}{*}{\textbf{Overall Avg.}} \\
\cmidrule(lr){2-5}\cmidrule(lr){6-9}\cmidrule(lr){10-13}\cmidrule(lr){14-17}
 & \textbf{TSO} & \textbf{MPDR} & \textbf{ASII} & \textbf{Avg.} &
 \textbf{AGBI} & \textbf{AGAR} & \textbf{AGLT} & \textbf{Avg.} &
 \textbf{FAM} & \textbf{SVA} & \textbf{FOM} & \textbf{Avg.} &
 \textbf{MEGL} & \textbf{AC} & \textbf{RLPC} & \textbf{Avg.} & \\
\midrule
GPT-4.1~\cite{openai_gpt41_2025}               & 18.9 & 29.7 & 27.7 & 25.4 & 78.0 & 59.1 & 61.0 & 66.0 & 30.4 & 38.9 & 41.9 & 37.1 & 25.5 & 24.3 & 37.3 & 29.0 & 39.4 \\
Gemini-2.5-Flash~\cite{comanici2025gemini}          & 28.6 & 35.8 & 24.8 & 29.7 & 75.2 & 71.4 & 63.1 & 69.9 & 29.2 & 31.3 & 44.2 & 34.9 & 22.6 & 26.6 & 31.2 & 26.8 & 40.3 \\
\midrule
Qwen2.5-VL-72B~\cite{qwen25vl_72b}        & 10.4 & \textbf{42.6} & 27.8 & 26.9 & 74.4 & 76.1 & 62.2 & 70.9 & 25.6 & \textbf{50.6} & 33.5 & 36.6 & 18.1 & 23.0 & 32.0 & 24.4 & 39.7 \\
Gemma-3-27B~\cite{gemma3_27b_2025}             & 38.4 & 42.0 & 15.7 & 32.0 & 69.0 & 50.5 & 55.6 & 58.4 & 21.8 & 14.3 & 28.4 & 21.5 & 15.7 & \textbf{29.0} & 25.7 & 23.5 & 33.8 \\
LLaMA-4-Scout-17B~\cite{llama4_scout_17b_2025} & 6.2  & 30.0 & 20.1 & 18.8 & 64.7 & 51.6 & 55.6 & 57.3 & 19.3 & 36.5 & 20.7 & 25.5 & 17.2 & 26.1 & 29.4 & 24.2 & 31.4 \\
InternVL3.5-14B~\cite{internvl35_8b_2025}     & \textbf{43.9} & 38.8 & \textbf{30.3} & 37.7 & 75.9 & 69.4 & 62.6 & 69.3 & 26.8 & 22.8 & 43.8 & 31.1 & 25.3 & 20.8 & \textbf{37.3} & 27.8 & 41.5 \\
Ovis-2.5-9B~\cite{ovis25_9b_2025}              & 0.1  & 30.6 & 26.0 & 18.9 & \textbf{79.7} & \textbf{76.2} & \textbf{64.7} & \textbf{73.5} & \textbf{33.6} & 57.2 & \textbf{49.6} & \textbf{46.8} & 27.5 & 23.4 & 36.7 & 29.2 & \textbf{42.1} \\
InternVL3.5-8B~\cite{internvl35_8b_2025}       & 41.3 & 31.3 & 28.1 & 33.6 & 77.6 & 71.6 & 61.4 & 70.2 & 26.3 & 21.2 & 41.5 & 29.7 & \textbf{33.1} & 21.2 & 38.1 & \textbf{30.8} & 41.1 \\
LLaVA-OneVision1.5-8B~\cite{llava_onevision15_8b_2025} & 26.6 & 28.7 & 23.0 & 26.1 & 76.8 & 67.5 & 58.8 & 67.7 & 29.8 & 33.9 & 37.1 & 33.6 & 25.2 & 17.6 & 31.9 & 24.9 & 38.1 \\
Qwen3-VL-8B~\cite{bai2025qwen3vl}             & 2.2  & 28.7 & 26.0 & 19.0 & 74.6 & 69.6 & 61.9 & 68.7 & 30.0 & 51.3 & 40.4 & 40.6 & 18.8 & 21.8 & 34.9 & 25.2 & 38.3 \\
MiniCPM-V4.5-8B~\cite{minicpm_v45_8b_2025}     & 19.1 & 26.3 & 26.0 & 23.8 & 77.9 & 72.3 & 63.2 & 71.1 & 30.2 & 43.7 & 45.2 & 39.7 & 24.1 & 22.9 & 27.9 & 24.9 & 39.9 \\
Qwen2.5-VL-7B~\cite{qwen25vl_72b}              & 14.6 & 28.0 & 22.9 & 21.8 & 69.7 & 59.3 & 52.9 & 60.6 & 33.0 & 36.0 & 47.1 & 38.7 & 21.1 & 20.3 & 26.3 & 22.6 & 35.9 \\
LLaVA-OneVision-7B~\cite{llava_onevision_7b_2024} & 33.3 & 24.9 & 23.7 & 27.3 & 67.1 & 58.0 & 52.3 & 59.1 & 28.9 & 22.4 & 38.9 & 30.1 & 22.8 & 22.4 & 32.8 & 26.0 & 35.6 \\
\midrule
\textbf{Avg.}          & 23.1 & 31.9 & 25.5 & 26.8 & 74.5 & 66.3 & 59.7 & 66.8 & 28.4 & 35.2 & 40.9 & 34.8 & 23.2 & 23.0 & 32.7 & 26.3 & \textbf{38.2} \\
\bottomrule
\end{tabular}
}
\end{adjustbox}
\caption{\textbf{HERBench results}. We report per-task accuracy (\%) for 13 leading MLLMs. The highest performance in each task column is marked in \textbf{bold}. The 4 largest-size models (first rows) were run on a representative 10\% subset ($\sim$2.6K questions), while the remaining models were evaluated on the full benchmark.}
\label{tab:main_results}
\end{table*}

To validate the challenges posed by \textbf{HERBench}, we conduct a comprehensive evaluation of current state-of-the-art Multimodal Large Language Models. Our experiments are designed to quantify their performance on tasks explicitly requiring the integration of multiple, temporally dispersed visual cues.

\paragraph{Setup.}
We evaluate 13 prominent MLLMs, including both closed- and open-source systems (Table~\ref{tab:main_results}). To isolate the effects of evidence aggregation from frame selection, all models receive the same budget of 16 uniformly sampled frames from each video.


\paragraph{Results.}
As shown in Table~\ref{tab:main_results}, performance is systematically poor. The mean accuracy across all 13 state-of-the-art models is 38.2\%, with the best model (Ovis-2.5-9B \cite{ovis25_9b_2025}) reaching only 42.1\% and the lowest (LLaMA-4-Scout-17B \cite{llama4_scout_17b_2025}) at 31.4\%. This narrow performance band, just 11--22 percentage above the 20\% random baseline, reveals that failure to integrate dispersed evidence is a pervasive limitation across all current architectures. 

The performance breakdown by task is telling. Models show relative competence on single-entity tracking tasks like \textit{[AGBI]} and \textit{[AGAR]} (Ovis-2.5-9B: 79.7\%, 76.2\%). This suggests they can track a single described entity. However, performance collapses on all tasks strictly requiring multi-cue aggregation. For \textit{[AC]} and \textit{[MEGL]}, mean accuracies are 23.0\% and 23.2\% respectively, barely above chance. Similarly, models fail at temporal ordering (\textit{[TSO]}), with scores as low as 0.1\%, demonstrating a clear inability to compose dispersed information.

In summary, these results demonstrate that while state-of-the-art MLLMs can track single entities, they fundamentally fail at the core challenge of multi-evidence compositional reasoning. Our controlled-frame evaluation confirms this deficit stems from a failure to \textit{integrate} information, not merely a failure to \textit{access} it.
\section{Analysis}
\label{sec:analysis}

This section analyzes the two major challenges highlighted by \textit{HERBench}: 
(\textbf{Q1}) how frame selection strategies affect performance through evidence retrieval, 
and (\textbf{Q2}) whether models can effectively aggregate evidence across the correct frames once retrieval uncertainty is removed.

\subsection{Isolating the Evidence Retrieval Bottleneck}\label{sec:analysis:frame_selection}

\paragraph{Frame Selection methods.}
To address \textbf{(Q1)}, i.e. the impact of evidence retrieval, we compare five strategies (all operating in the same BLIP \cite{li2022blip} embedding space for fairness): \textbf{AKS} \cite{tang2025aks} learns a keyframe policy that balances relevance and temporal coverage; \textbf{BOLT-ITS} \cite{liu2025bolt} using inverse transform sampling to select query-relevant frames; \textbf{Uniform} takes evenly spaced frames; \textbf{Vanilla-BLIP} retrieves frames with highest cosine similarity to the question; \textbf{Oracle Frames (OF)} use frame indices gathered from our benchmark’s construction pipeline along with a few non-evidence frames to match the fixed frame budget (they are meant to represent a best-case retrieval setting rather than the oracle-only fusion setting studied in Sec.~\ref{sec:analysis:aggregation}). They are applied only in tasks where relevant evidence is scarce or confined to very short portions of the video (notably \textit{[TSO]}, \textit{[FAM]}, \textit{[SVA]}).


\begin{table}[t]
\centering
\footnotesize 
\setlength{\tabcolsep}{3pt} 
\begin{tabular}{lccc}
\toprule
\textbf{Frame Selection} & \textbf{InternVL3.5-14B} & \textbf{Qwen3-VL-8B} & \textbf{Ovis-2.5-9B} \\
\midrule
Uniform          & 42.7 & 37.7 & 43.1 \\
Vanilla-BLIP     & 42.1 & 37.9 & 41.6 \\
BOLT-ITS         & 41.1 & 38.4 & 42.1 \\
AKS              & 42.7 & 36.2 & 42.6 \\
Oracle Frames (OF)  & \textbf{47.8} & \textbf{41.0} & \textbf{47.9} \\
\bottomrule
\end{tabular}
\caption{\textbf{Mean accuracy by frame selection method and model.} Rows list frame selection methods. Each cell shows the mean accuracy over 1200 questions (100 from each task).}
\label{tab:frame-selection-means}
\end{table}

\paragraph{Performance across frame selection strategies.}
Table~\ref{tab:frame-selection-means} presents the accuracy, averaged across all tasks, for each selection method applied to three representative models. Learned selectors such as AKS and BOLT-ITS outperform simple uniform sampling on many tasks, yet still trail behind the Oracle Frames (OF) configuration, a performance gap that is even more pronounced in the per-task breakdown (see Supplementary), reinforcing the fact that evidence retrieval remains a major performance bottleneck. More importantly, even when the model is provided with the correct evidence frames (using OF), performance gains are limited, with accuracy remaining below 50\%,  indicating that access to the right information alone is insufficient for successful multi-evidence reasoning. This finding aligns with our broader observation that current models underweight or fail to integrate critical cues, even when ground-truth evidence is fully available.

\subsection{Evidence Aggregation with Oracle-Only Frames}
\label{sec:analysis:aggregation}

Having established that evidence retrieval is a significant bottleneck (Sec.~\ref{sec:analysis:frame_selection}), we now turn to \textbf{(Q2)}: can models effectively aggregate evidence even when retrieval uncertainty is removed? To isolate the fusion capability from the retrieval challenge, we conduct a targeted study on a subset of HERBench supplying models with \textbf{only the manually curated ground-truth frames} (the "oracle" set). In a parallel human study under the same oracle presentation format, annotators achieved 95.7\% accuracy, indicating that the curated oracle frame-set is generally sufficient for resolving the question.

\begin{figure}[t]
  \centering
  \includegraphics[width=0.49\textwidth]{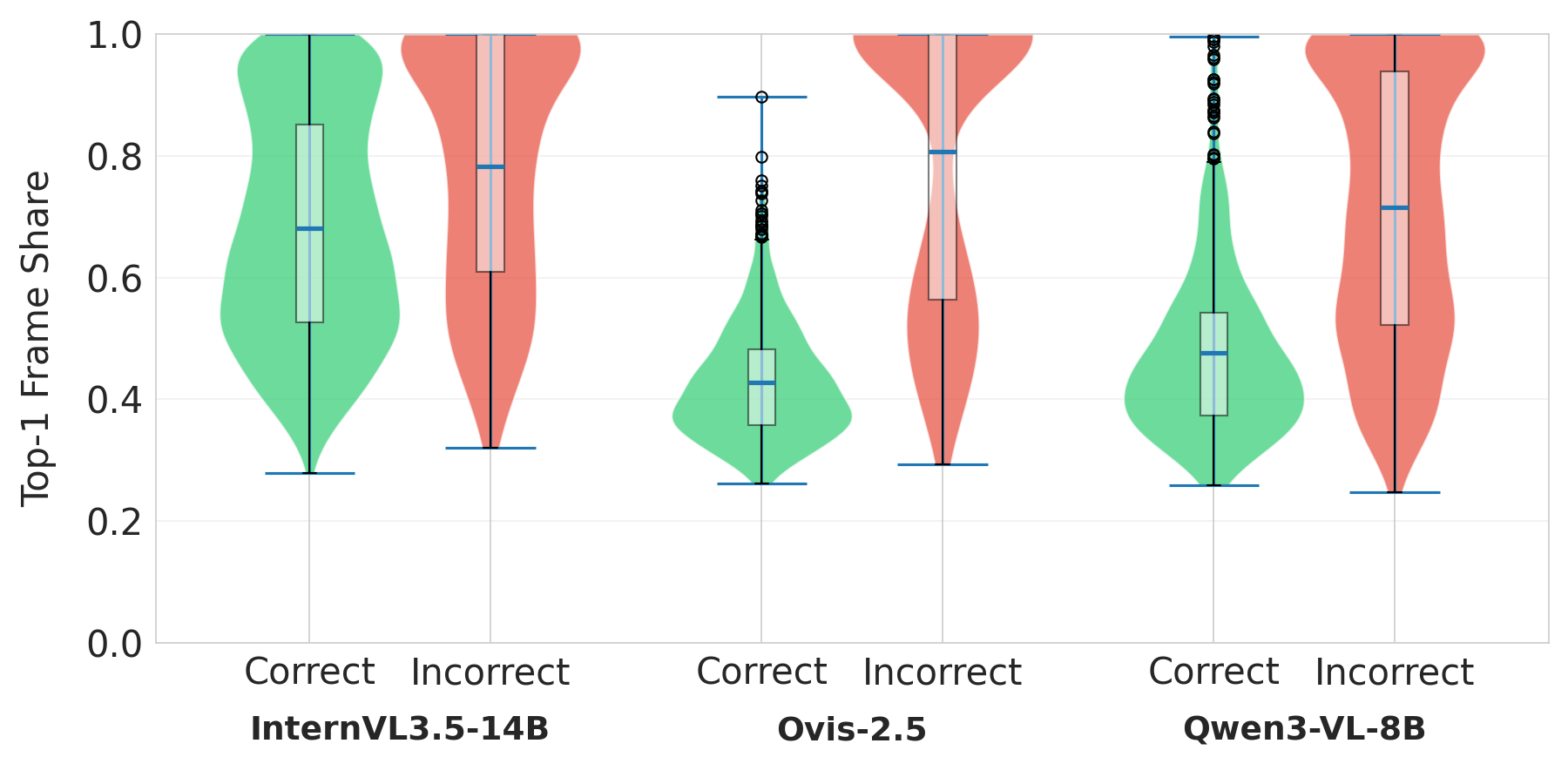}
  \caption{\textbf{Top-1 frame share under oracle-only frames.} We plot the maximum normalized frame-importance share for three models, separated by correct and incorrect predictions. Correct answers distribute importance more evenly, whereas errors concentrate it on a single frame, indicating weak multi-evidence fusion even when evidence-bearing frames are provided.}
  \label{fig:top1-share}
\end{figure}

\paragraph{Measuring Frame-Level Contribution.}
For each item, we compute per-frame \emph{deltas} and \emph{shares} that quantify how much each frame contributes to the model’s \emph{own} predicted option:
\begin{enumerate}
    \item \textbf{Full prediction.} Run the model on all oracle frames and compute $\log p_{\text{full}}$, where $p$ is the \emph{post-softmax} probability of the chosen \emph{letter token} (\texttt{A}--\texttt{E}), with the softmax taken only over these candidate tokens.
    \item \textbf{Leave-one-out re-run.} For each frame $i$, re-run with that frame excluded (the context contains the remaining $n-1$ frames) to obtain $\log p_{\text{minus}[i]}$.
    \item \textbf{Delta.} $\Delta_i = \log p_{\text{full}} - \log p_{\text{minus}[i]}$; positive $\Delta_i$ means frame $i$ supports the model’s chosen option.
    \item \textbf{Share.} $s_i = \Delta_i^+ / \sum_j \Delta_j^+$, yielding a normalized importance distribution across frames.
\end{enumerate}

\paragraph{Diagnosing Fusion Failures via Importance Distribution.}

We analyze per-frame importance shares to understand why models succeed or fail under oracle-only inputs, summarizing each item using the \textbf{Top-1 Share} ($\max_i s_i$) as shown in Figure~\ref{fig:top1-share}. This statistic captures how strongly a model concentrates its decision on a single frame. The distributions reveal a consistent pattern: correct predictions (green) exhibit substantially more balanced allocations, with mean Top-1 shares near 0.5. Incorrect predictions (red), in contrast, show pronounced over-concentration, with Top-1 shares frequently approaching 0.8. This separation indicates that errors arise not merely from insufficient signal, but from \emph{misallocation} of attention-models place disproportionate weight on one frame while failing to assign sufficient importance to the multiple, distributed evidential cues present across the oracle set. Because HERBench questions structurally require multi-frame reasoning, this behavior shows that \textbf{the fusion module itself-independent of retrieval—remains a primary source of failure}.
\section{Conclusion}
\label{sec:Conclusions}
We introduced HERBench, a VideoQA benchmark comprising 26,806 questions across 12 tasks, each designed to structurally enforce the aggregation of $k \geq 3$ distinct, temporally separated visual cues. To quantify evidential demand, we proposed the Minimum Required Frame-Set (MRFS) metric; under a canonical evaluation protocol, HERBench attains the highest mean MRFS among the benchmarks considered. Experiments on 13 state-of-the-art MLLMs reveal a narrow accuracy range (31.4--42.1\%), only modestly above the 20\% chance baseline, exposing persistent limitations in multi-evidence reasoning. We trace these failures to two main bottlenecks: a \emph{retrieval} deficit, where frame selectors fail to recover all necessary cues, and a \emph{fusion} deficit, where models fail to combine them even when available, often collapsing onto a single frame. HERBench therefore establishes a principled benchmark for studying high-evidential-requirement video understanding, exposes substantial headroom beyond single-cue success, and points toward future improvements in retrieval-aware querying and distributed evidence fusion.

{\small
\bibliographystyle{IEEEtranN}
\bibliography{main}

@String(CVPR= {IEEE Conf. Comput. Vis. Pattern Recog.})

@String(ICCV= {Int. Conf. Comput. Vis.})

@String(ECCV= {Eur. Conf. Comput. Vis.})

@String(NIPS= {Adv. Neural Inform. Process. Syst.})

@String(ICIP = {IEEE Int. Conf. Image Process.})

@String(ICLR = {Int. Conf. Learn. Represent.})

@String(CVPR  = {CVPR})

@String(ICCV  = {ICCV})

@String(ECCV  = {ECCV})

@String(NIPS  = {NeurIPS})

@String(ICIP  = {ICIP})

@String(ICLR  = {ICLR})

@inproceedings{lei2018tvqa,
  author    = {Jie Lei and Licheng Yu and Mohit Bansal and Tamara L. Berg},
  title     = {TVQA: Localized, Compositional Video Question Answering},
  booktitle = {EMNLP},
  year      = {2018},
  note      = {arXiv:1809.01696}
}

@inproceedings{xiao2021nextqa,
  author    = {Junbin Xiao and Xindi Shang and Angela Yao and Tat{-}Seng Chua},
  title     = {{NExT{-}QA}: Next Phase of Question-Answering to Explaining Temporal Actions},
  booktitle = CVPR,
  year      = {2021},
  url       = {https://arxiv.org/abs/2105.08276}
}

@inproceedings{li2024mvbench,
  author    = {Kunchang Li and Yali Wang and Yinan He and Yizhuo Li and Yi Wang and Yi Liu and Zun Wang and Jilan Xu and Guo Chen and Ping Luo and Limin Wang and Yu Qiao},
  title     = {{MVBench}: A Comprehensive Multi-modal Video Understanding Benchmark},
  booktitle = CVPR,
  year      = {2024},
  url       = {https://openaccess.thecvf.com/content/CVPR2024/papers/Li_MVBench_A_Comprehensive_Multi-modal_Video_Understanding_Benchmark_CVPR_2024_paper.pdf}
}

@inproceedings{longvideobench2024,
  author    = {Haoning Wu and Dongxu Li and Bei Chen and Junnan Li},
  title     = {LongVideoBench: A Benchmark for Long-context Interleaved Video-Language Understanding},
  booktitle = NIPS,
  year      = {2024},
  url       = {https://proceedings.neurips.cc/paper_files/paper/2024/hash/329ad516cf7a6ac306f29882e9c77558-Abstract-Datasets_and_Benchmarks_Track.html},
  note      = {arXiv:2407.15754}
}

@article{mangalam2023egoschema,
  title={Egoschema: A diagnostic benchmark for very long-form video language understanding},
  author={Mangalam, Karttikeya and Akshulakov, Raiymbek and Malik, Jitendra},
  journal={Advances in Neural Information Processing Systems},
  volume={36},
  pages={46212--46244},
  year={2023}
}

@inproceedings{mmbenchvideo2024,
  author    = {Xinyu Fang and Kangrui Mao and Haodong Duan and Xiangyu Zhao and Yining Li and Dahua Lin and Kai Chen},
  title     = {{MMBench{-}Video}: A Long-Form Multi-Shot Benchmark for Holistic Video Understanding},
  booktitle = NIPS,
  year      = {2024},
  note      = {Datasets and Benchmarks Track, arXiv:2406.14515}
}

@inproceedings{fu2024videomme,
  author    = {Fu, Chaoyou and Dai, Yuhan and Luo, Yongdong and Li, Lei and Ren, Shuhuai and Zhang, Renrui and Wang, Zihan and Zhou, Chenyu and Shen, Yunhang and Zhang, Mengdan and Chen, Peixian and Li, Yanwei and Lin, Shaohui and Zhao, Sirui and Li, Ke and Xu, Tong and Zheng, Xiawu and Chen, Enhong and Shan, Caifeng and He, Ran and Sun, Xing},
    title     = {Video-MME: The First-Ever Comprehensive Evaluation Benchmark of Multi-modal LLMs in Video Analysis},
    booktitle = {Proceedings of the IEEE/CVF Conference on Computer Vision and Pattern Recognition (CVPR)},
    month     = {June},
    year      = {2025},
    pages     = {24108-24118}
}

@inproceedings{xiao2021star,
  author    = {Bo Wu and Shoubin Yu and Zhenfang Chen and Joshua B. Tenenbaum and Chuang Gan},
  title     = {{STAR}: A Benchmark for Situated Reasoning in Real-World Videos},
  booktitle = NIPS,
  year      = {2021},
  note      = {Datasets \& Benchmarks Track},
  url       = {https://arxiv.org/abs/2405.09711}
}

@inproceedings{tempcompass2024,
  author    = {Yuanxin Liu and Shicheng Li and Yi Liu and Yuxiang Wang and Shuhuai Ren and Lei Li and Sishuo Chen and Xu Sun and Lu Hou},
  title     = {TempCompass: Do Video LLMs Really Understand Videos?},
  booktitle = {Findings of ACL},
  year      = {2024},
  url       = {https://aclanthology.org/2024.findings-acl.517/},
  note      = {arXiv:2403.00476}
}

@inproceedings{tang2025aks,
  author    = {Xi Tang and Jihao Qiu and Lingxi Xie and Yunjie Tian and Jianbin Jiao and Qixiang Ye},
  title     = {Adaptive Keyframe Sampling for Long Video Understanding},
  booktitle = CVPR,
  year      = {2025},
  note      = {arXiv:2502.21271}
}

@inproceedings{tapaswi2016movieqa,
  author    = {Makarand Tapaswi and Yukun Zhu and Rainer Stiefelhagen and Antonio Torralba and Raquel Urtasun and Sanja Fidler},
  title     = {MovieQA: Understanding Stories in Movies through Question-Answering},
  booktitle = CVPR,
  year      = {2016},
  url       = {https://openaccess.thecvf.com/content_cvpr_2016/papers/Tapaswi_MovieQA_Understanding_Stories_CVPR_2016_paper.pdf}
}

@inproceedings{jang2017tgif,
  title={Tgif-qa: Toward spatio-temporal reasoning in visual question answering},
  author={Jang, Yunseok and Song, Yale and Yu, Youngjae and Kim, Youngjin and Kim, Gunhee},
  booktitle={Proceedings of the IEEE conference on computer vision and pattern recognition},
  pages={2758--2766},
  year={2017}
}

@inproceedings{grundemclaughlin2021agqa,
  author    = {Madeleine Grunde{-}McLaughlin and Ranjay Krishna and Maneesh Agrawala},
  title     = {AGQA: A Benchmark for Compositional Spatio-Temporal Reasoning},
  booktitle = CVPR,
  year      = {2021},
  url       = {https://openaccess.thecvf.com/content/CVPR2021/papers/Grunde-McLaughlin_AGQA_A_Benchmark_for_Compositional_Spatio-Temporal_Reasoning_CVPR_2021_paper.pdf}
}

@inproceedings{yi2020clevrer,
  author    = {Kexin Yi and Chuang Gan and Yunzhu Li and Pushmeet Kohli and Jiajun Wu and Antonio Torralba and Joshua B. Tenenbaum},
  title     = {CLEVRER: Collision Events for Video Representation and Reasoning},
  booktitle = ICLR,
  year      = {2020},
  url       = {https://arxiv.org/abs/1910.01442}
}

@inproceedings{girdhar2020cater,
  author    = {Rohit Girdhar and Deva Ramanan},
  title     = {CATER: A Diagnostic Dataset for Compositional Actions \& Temporal Reasoning},
  booktitle = ICLR,
  year      = {2020},
  url       = {https://arxiv.org/abs/1910.04744}
}

@inproceedings{xiao2024visground,
  author    = {Junbin Xiao and Angela Yao and Yicong Li and Tat{-}Seng Chua},
  title     = {Can I Trust Your Answer? Visually Grounded Video Question Answering},
  booktitle = CVPR,
  year      = {2024},
  url       = {https://openaccess.thecvf.com/content/CVPR2024/papers/Xiao_Can_I_Trust_Your_Answer_Visually_Grounded_Video_Question_Answering_CVPR_2024_paper.pdf}
}

@inproceedings{wojke2017simple,
  title={Simple Online and Realtime Tracking with a Deep Association Metric},
  author={Wojke, Nicolai and Bewley, Alex and Paulus, Dietrich},
  booktitle={2017 IEEE International Conference on Image Processing (ICIP)},
  year={2017}
}

@misc{openai2024gpt4o,
  title={GPT-4o},
  author={{OpenAI}},
  year={2024},
  note={Model card and system card}
}

@misc{qwen2,
  title={Qwen2: A family of open large language models},
  author={{Qwen Team}},
  year={2024},
  note={Alibaba Cloud}
}

@misc{qwen2_5,
  title={Qwen2.5 Technical Report},
  author={{Qwen Team}},
  year={2024},
  note={Alibaba Cloud}
}

@misc{llama3,
  title={The Llama 3 Herd of Models},
  author={{Meta AI}},
  year={2024},
  note={Model release report}
}

@misc{vicuna15,
  title={Vicuna v1.5: An Open-Source Chatbot},
  author={Chiang, Wei-Lin and Li, Zhuohan and others},
  year={2023},
  note={FastChat project report}
}

@InProceedings{liu2025bolt,
    author    = {Liu, Shuming and Zhao, Chen and Xu, Tianqi and Ghanem, Bernard},
    title     = {BOLT: Boost Large Vision-Language Model Without Training for Long-form Video Understanding},
    booktitle = {Proceedings of the IEEE/CVF Conference on Computer Vision and Pattern Recognition (CVPR)},
    month     = {June},
    year      = {2025},
    pages     = {3318-3327}
}

@InProceedings{Wu_2022_CVPR,
  author    = {Wu, Jiannan and Jiang, Yi and Sun, Peize and Yuan, Zehuan and Luo, Ping},
  title     = {Language as Queries for Referring Video Object Segmentation},
  booktitle = {Proceedings of the IEEE/CVF Conference on Computer Vision and Pattern Recognition (CVPR)},
  year      = {2022}
}

@InProceedings{Botach_2022_CVPR,
  author    = {Botach, Adam and Zheltonozhskii, Evgenii and Baskin, Chaim},
  title     = {End-to-End Referring Video Object Segmentation With Multimodal Transformers},
  booktitle = {Proceedings of the IEEE/CVF Conference on Computer Vision and Pattern Recognition (CVPR)},
  year      = {2022}
}

@InProceedings{Gavrilyuk_2018_CVPR,
  author    = {Gavrilyuk, Kirill and Ghodrati, Amir and Li, Zhenyang and Snoek, Cees G. M.},
  title     = {Actor and Action Video Segmentation From a Sentence},
  booktitle = {Proceedings of the IEEE Conference on Computer Vision and Pattern Recognition (CVPR)},
  year      = {2018}
}

@InProceedings{Seo_2020_ECCV,
  author    = {Seo, Seonguk and Lee, Joon-Young and Han, Bohyung},
  title     = {URVOS: Unified Referring Video Object Segmentation Network with a Large-Scale Benchmark},
  booktitle = {European Conference on Computer Vision (ECCV)},
  year      = {2020}
}

@inproceedings{li2022blip,
  title={Blip: Bootstrapping language-image pre-training for unified vision-language understanding and generation},
  author={Li, Junnan and Li, Dongxu and Xiong, Caiming and Hoi, Steven},
  booktitle={International conference on machine learning},
  pages={12888--12900},
  year={2022},
  organization={PMLR}
}

@inproceedings{perrett2025hdepic,
  author={Perrett, Toby and Darkhalil, Ahmad and Sinha, Saptarshi and Emara, Omar and Pollard, Sam and Parida, Kranti Kumar and Liu, Kaiting and Gatti, Prajwal and Bansal, Siddhant and Flanagan, Kevin and Chalk, Jacob and Zhu, Zhifan and Guerrier, Rhodri and Abdelazim, Fahd and Zhu, Bin and Moltisanti, Davide and Wray, Michael and Doughty, Hazel and Damen, Dima},
  booktitle={2025 IEEE/CVF Conference on Computer Vision and Pattern Recognition (CVPR)}, 
  title={HD-EPIC: A Highly-Detailed Egocentric Video Dataset}, 
  year={2025},
  volume={},
  number={},
  pages={23901-23913},
  doi={10.1109/CVPR52734.2025.02226}}

@inproceedings{wildtrack,
  title={Wildtrack: A multi-camera hd dataset for dense unscripted pedestrian detection},
  author={Chavdarova, Tatjana and Baqu{\'e}, Pierre and Bouquet, St{\'e}phane and Maksai, Andrii and Jose, Cijo and Bagautdinov, Timur and Lettry, Louis and Fua, Pascal and Van Gool, Luc and Fleuret, Fran{\c{c}}ois},
  booktitle={Proceedings of the IEEE conference on computer vision and pattern recognition},
  pages={5030--5039},
  year={2018}
}

@inproceedings{personpath22,
  title={Large scale real-world multi-person tracking},
  author={Shuai, Bing and Bergamo, Alessandro and Buechler, Uta and Berneshawi, Andrew and Boden, Alyssa and Tighe, Joseph},
  booktitle={European Conference on Computer Vision},
  pages={504--521},
  year={2022},
  organization={Springer}
}

@misc{openai_gpt41_2025,
  title        = {Introducing GPT-4.1 in the API},
  author       = {{OpenAI}},
  year         = {2025},
  howpublished = {\url{https://openai.com/index/gpt-4-1/}},
}

@article{comanici2025gemini,
  title={Gemini 2.5: Pushing the frontier with advanced reasoning, multimodality, long context, and next generation agentic capabilities},
  author={Comanici, Gheorghe and Bieber, Eric and Schaekermann, Mike and Pasupat, Ice and Sachdeva, Noveen and Dhillon, Inderjit and Blistein, Marcel and Ram, Ori and Zhang, Dan and Rosen, Evan and others},
  journal={arXiv preprint arXiv:2507.06261},
  year={2025}
}

@article{qwen25vl_72b,
  title   = {Qwen2.5-VL Technical Report},
  author  = {Bai, Shuai and others},
  journal = {arXiv preprint arXiv:2502.13923},
  year    = {2025},
  url     = {https://arxiv.org/abs/2502.13923}
}

@article{gemma3_27b_2025,
  title   = {Gemma 3 Technical Report},
  author  = {{Gemma Team}},
  journal = {arXiv preprint arXiv:2503.19786},
  year    = {2025},
  url     = {https://arxiv.org/abs/2503.19786}
}

@misc{llama4_scout_17b_2025,
  title        = {Llama 4 Model Card (Scout Models)},
  author       = {{Meta AI}},
  year         = {2025},
  howpublished = {\url{https://github.com/meta-llama/llama-models/blob/main/models/llama4/MODEL_CARD.md}},
  note         = {Model card; no public technical report for Llama~4 Scout as of Nov 2025}
}

@article{ovis25_9b_2025,
  title   = {Ovis2.5 Technical Report},
  author  = {Lu, Shiyin and others},
  journal = {arXiv preprint arXiv:2508.11737},
  year    = {2025},
  url     = {https://arxiv.org/abs/2508.11737}
}

@article{internvl35_8b_2025,
  title   = {InternVL3.5: Advancing Open-Source Multimodal Models in Versatility, Reasoning, and Efficiency},
  author  = {Wang, Weiyun and others},
  journal = {arXiv preprint arXiv:2508.18265},
  year    = {2025},
  url     = {https://arxiv.org/abs/2508.18265}
}

@article{llava_onevision15_8b_2025,
  title   = {LLaVA-OneVision-1.5: Fully Open Framework for Democratized Multimodal Training},
  author  = {An, Xiang and others},
  journal = {arXiv preprint arXiv:2509.23661},
  year    = {2025},
  url     = {https://arxiv.org/abs/2509.23661}
}

@article{minicpm_v45_8b_2025,
  title   = {MiniCPM-V 4.5: Cooking Efficient MLLMs via Architecture, Data, and Training Recipe},
  author  = {Yu, Tianyu and others},
  journal = {arXiv preprint arXiv:2509.18154},
  year    = {2025},
  url     = {https://arxiv.org/abs/2509.18154}
}

@article{llava_onevision_7b_2024,
  title   = {LLaVA-OneVision: Easy Visual Task Transfer},
  author  = {Li, Bo and others},
  journal = {arXiv preprint arXiv:2408.03326},
  year    = {2024},
  url     = {https://arxiv.org/abs/2408.03326}
}

@INPROCEEDINGS{xu2016msrvtt,
  author={Xu, Jun and Mei, Tao and Yao, Ting and Rui, Yong},
  booktitle={2016 IEEE Conference on Computer Vision and Pattern Recognition (CVPR)}, 
  title={MSR-VTT: A Large Video Description Dataset for Bridging Video and Language}, 
  year={2016},
  volume={},
  number={},
  pages={5288-5296},
  doi={10.1109/CVPR.2016.571}}

@inproceedings{xu2017video,
  title={Video Question Answering via Gradually Refined Attention over Appearance and Motion},
  author={Xu, Dejing and Zhao, Zhou and Xiao, Jun and Wu, Fei and Zhang, Hanwang and He, Xiangnan and Zhuang, Yueting},
  booktitle={ACM Multimedia},
  year={2017}
}

@inproceedings{breaking-down,
title = {Breaking Down Video LLM Benchmarks: Knowledge, Spatial Perception, or True Temporal Understanding?},
booktitle = {NeurIPS Workshop},
author={Feng, Bo and Lai, Zhengfeng and Li, Shiyu and Wang, Zizhen and Wang, Simon and Huang, Ping and Cao, Meng},
year = {2025},
URL = {https://arxiv.org/abs/2505.14321}
}

@article{damonlpsg2023videollama,
  author = {Zhang, Hang and Li, Xin and Bing, Lidong},
  title = {Video-LLaMA: An Instruction-tuned Audio-Visual Language Model for Video Understanding},
  year = 2023,
  journal = {arXiv preprint arXiv:2306.02858},
  url = {https://arxiv.org/abs/2306.02858}
}

@inproceedings{Maaz2023VideoChatGPT,
    title={Video-ChatGPT: Towards Detailed Video Understanding via Large Vision and Language Models},
    author={Maaz, Muhammad and Rasheed, Hanoona and Khan, Salman and Khan, Fahad Shahbaz},
    booktitle={Proceedings of the 62nd Annual Meeting of the Association for Computational Linguistics (ACL 2024)},
    year={2024}
}

@inproceedings{li2023blip2,
    author = {Li, Junnan and Li, Dongxu and Savarese, Silvio and Hoi, Steven},
    title = {BLIP-2: bootstrapping language-image pre-training with frozen image encoders and large language models},
    year = {2023},
    publisher = {JMLR.org},
    booktitle = {Proceedings of the 40th International Conference on Machine Learning},
    articleno = {814},
    numpages = {13},
    location = {Honolulu, Hawaii, USA},
    series = {ICML'23}
}

@inproceedings{dai2023instructblip,
    author = {Dai, Wenliang and Li, Junnan and Li, Dongxu and Tiong, Anthony Meng Huat and Zhao, Junqi and Wang, Weisheng and Li, Boyang and Fung, Pascale and Hoi, Steven},
    title = {InstructBLIP: towards general-purpose vision-language models with instruction tuning},
    year = {2023},
    publisher = {Curran Associates Inc.},
    address = {Red Hook, NY, USA},
    booktitle = {Proceedings of the 37th International Conference on Neural Information Processing Systems},
    articleno = {2142},
    numpages = {18},
    location = {New Orleans, LA, USA},
    series = {NIPS '23}
}

@InProceedings{nagrani2025minerva,
    author    = {Nagrani, Arsha and Menon, Sachit and Iscen, Ahmet and Buch, Shyamal and Mehran, Ramin and Jha, Nilpa and Hauth, Anja and Zhu, Yukun and Vondrick, Carl and Sirotenko, Mikhail and Schmid, Cordelia and Weyand, Tobias},
    title     = {MINERVA: Evaluating Complex Video Reasoning},
    booktitle = {Proceedings of the IEEE/CVF International Conference on Computer Vision (ICCV)},
    month     = {October},
    year      = {2025},
    pages     = {23968-23978}
}

@article{robinson2025rfdetr,
  title={RF-DETR: neural architecture search for real-time detection transformers},
  author={Robinson, Isaac and Robicheaux, Peter and Popov, Matvei and Ramanan, Deva and Peri, Neehar},
  journal={arXiv preprint arXiv:2511.09554},
  year={2025}
}

@software{yolo11_ultralytics,
  author  = {Glenn Jocher and Jing Qiu},
  title   = {Ultralytics YOLO11},
  version = {11.0.0},
  year    = {2024},
  url     = {https://github.com/ultralytics/ultralytics}
}

@inproceedings{liu2024groundingdino,
  title={Grounding {DINO}: Marrying {DINO} with Grounded Pre-Training for Open-Set Object Detection},
  author={Liu, Shilong and Zeng, Zhaoyang and Ren, Tianhe and Li, Feng and Zhang, Hao and Yang, Jie and Jiang, Qing and Li, Chunyuan and Yang, Jianwei and Su, Hang and Zhu, Jun and Zhang, Lei},
  booktitle={European Conference on Computer Vision (ECCV)},
  pages={38--56},
  year={2024},
  publisher={Springer}
}

@inproceedings{zhang2022bytetrack,
  title={{ByteTrack}: Multi-Object Tracking by Associating Every Detection Box},
  author={Zhang, Yifu and Sun, Peize and Jiang, Yi and Yu, Dongdong and Weng, Fucheng and Yuan, Zehuan and Luo, Ping and Liu, Wenyu and Wang, Xinggang},
  booktitle={European Conference on Computer Vision (ECCV)},
  pages={1--21},
  year={2022},
  publisher={Springer}
}

@inproceedings{sun2022dancetrack,
  title={{DanceTrack}: Multi-Object Tracking in Uniform Appearance and Diverse Motion},
  author={Sun, Peize and Cao, Jinkun and Jiang, Yi and Yuan, Zehuan and Bai, Song and Kitani, Kris and Luo, Ping},
  booktitle={Proceedings of the IEEE/CVF Conference on Computer Vision and Pattern Recognition (CVPR)},
  pages={20993--21002},
  year={2022}
}

@article{soucek2020transnetv2,
  title={{TransNet V2}: An Effective Deep Network Architecture for Fast Shot Transition Detection},
  author={Sou\v{c}ek, Tom\'{a}\v{s} and Loko\v{c}, Jakub},
  journal={arXiv preprint arXiv:2008.04838},
  year={2020}
}

@inproceedings{lin2014coco,
  title={Microsoft {COCO}: Common Objects in Context},
  author={Lin, Tsung-Yi and Maire, Michael and Belongie, Serge and Hays, James and Perona, Pietro and Ramanan, Deva and Doll\'{a}r, Piotr and Zitnick, C. Lawrence},
  booktitle={European Conference on Computer Vision (ECCV)},
  pages={740--755},
  year={2014},
  publisher={Springer}
}

@article{bai2025qwen3vl,
  title={Qwen3-vl technical report},
  author={Bai, Shuai and Cai, Yuxuan and Chen, Ruizhe and Chen, Keqin and Chen, Xionghui and Cheng, Zesen and Deng, Lianghao and Ding, Wei and Gao, Chang and Ge, Chunjiang and others},
  journal={arXiv preprint arXiv:2511.21631},
  year={2025}
}

@inproceedings{ye2025re,
  title={Re-thinking temporal search for long-form video understanding},
  author={Ye, Jinhui and Wang, Zihan and Sun, Haosen and Chandrasegaran, Keshigeyan and Durante, Zane and Eyzaguirre, Cristobal and Bisk, Yonatan and Niebles, Juan Carlos and Adeli, Ehsan and Fei-Fei, Li and others},
  booktitle={Proceedings of the IEEE/CVF Conference on Computer Vision and Pattern Recognition},
  pages={8579--8591},
  year={2025}
}
}

\clearpage

\begin{center}
    {\LARGE\bfseries HERBench: A Benchmark for Multi-Evidence Integration in Video Question Answering\par}
    \vspace{0.6em}
    {\Large\bfseries Supplementary Material\par}
    \vspace{1.2em}
\end{center}

\renewcommand{\thesection}{\arabic{section}}
\renewcommand{\thesubsection}{\thesection.\arabic{subsection}}
\renewcommand{\thesubsubsection}{\thesubsection.\arabic{subsubsection}}

\setcounter{section}{6}
\setcounter{subsection}{0}
\setcounter{subsubsection}{0}


\section{Implementation Details}
\label{app:implementation}

This section provides comprehensive implementation details for the HERBench construction pipeline, which employs a tripartite structure to enforce high evidential requirements (ER). We detail the algorithms, mathematical formulations, thresholds, and quality control procedures used to transform raw videos into the final dataset.

\subsection{Track Ranking and Selection}
\label{app:track-ranking}


\paragraph{Tracking and Trajectory Refinement.}
To ensure robust performance in our benchmark's highly dynamic environments---characterized by dense crowds and significant depth variations---we conducted a rigorous empirical evaluation of state-of-the-art detection and tracking pipelines. Specifically, we benchmarked three leading detectors (\textit{RF-DETR-L}~\cite{robinson2025rfdetr}, \textit{YOLO-v11-x}~\cite{yolo11_ultralytics}, and \textit{Grounding-DINO}~\cite{liu2024groundingdino}) in combination with two widely adopted multi-object trackers (\textit{DeepSORT}~\cite{wojke2017simple} and \textit{ByteTrack}~\cite{zhang2022bytetrack}). This evaluation was performed on a curated subset of the DanceTrack~\cite{sun2022dancetrack} benchmark, explicitly selected to mirror the crowd density, complex motion dynamics, and high inter-subject appearance diversity typical of our video corpus.

As detailed in Table~\ref{tab:tracking_benchmark}, the combination of \textit{RF-DETR-L}~\cite{robinson2025rfdetr} and \textit{DeepSORT}~\cite{wojke2017simple} yielded the highest tracking fidelity, achieving a HOTA of 47.1\% and an IDF1 of 53.2\%. Consequently, we adopted this configuration as the foundation of our pipeline. We utilize this detection-tracking stack by applying a high-recall \textit{RF-DETR}~\cite{robinson2025rfdetr} detector with a confidence threshold of 0.3 and a per-frame cap of 300 detections. Association within \textit{DeepSORT}~\cite{wojke2017simple} uses a two-stage IoU matching: high-confidence detections (score $> 0.5$) are matched with an IoU threshold of 0.7, followed by lower-confidence detections with a relaxed IoU threshold of 0.35.

\begin{table}[h]
\centering
\resizebox{0.65\columnwidth}{!}{%
\begin{tabular}{l|ccc}
\toprule
\textbf{HOTA / IDF1 (\%)} & \textbf{RF-DETR-L} & \textbf{YOLO-v11-x} & \textbf{Grounding-DINO} \\
\midrule
\textbf{DeepSORT} & \textbf{47.1 / 53.2} & 43.3 / 51.1 & 42.9 / 50.6 \\
\textbf{ByteTracker} & 46.9 / 52.7 & 43.3 / 51.0 & 42.6 / 50.5 \\
\bottomrule
\end{tabular}%
}
\caption{Tracking pipeline benchmarking on a curated DanceTrack~\cite{sun2022dancetrack} subset mirroring HERBench's dynamic characteristics. The combination of RF-DETR-L and DeepSORT achieved the highest scores and was selected for our data generation pipeline.}
\label{tab:tracking_benchmark}
\end{table}

\begin{figure*}[!t] 
    \centering
    \includegraphics[width=\textwidth]{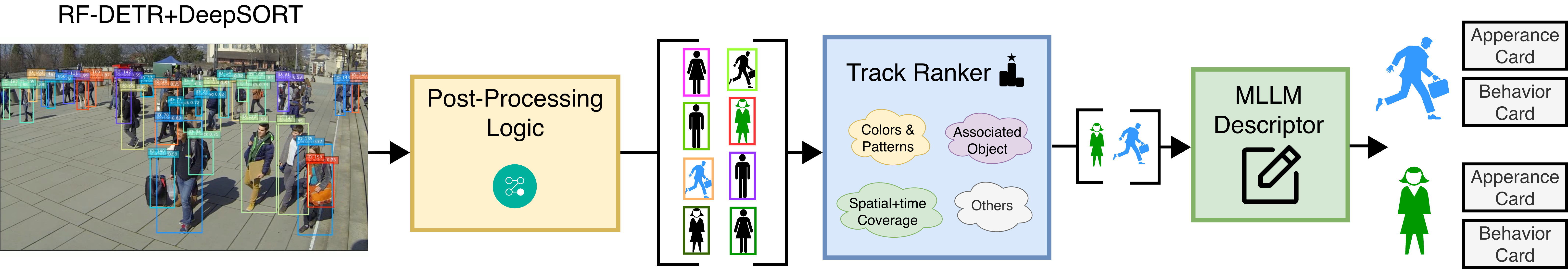} 
    \caption{\textbf{Tracking, post-processing, and ranking pipeline.} RF-DETR~\cite{robinson2025rfdetr} detections are linked with DeepSORT~\cite{wojke2017simple} into raw person tracks, followed by outlier removal, gap interpolation, and Gaussian smoothing. A TrackRanker then scores and selects salient trajectories, which are passed to an MLLM descriptor module to generate temporally decoupled appearance (A) and behavior (B) cards that serve as the scaffold for downstream HERBench tasks.}
    \label{fig:track_pipeline} 
\end{figure*}

To enforce physical plausibility, we apply an \textit{outlier removal} step that explicitly discards per-frame boxes implying implausible motion (velocity $> 50$ pixels/frame) to eliminate spurious detections. To ensure continuity, we apply gap interpolation for missing detections up to 30 frames (1s at 30 fps) and trajectory smoothing via Gaussian filtering (window size 5). We specifically address track fragmentation by detecting merge candidates $(T_i, T_j)$ that are temporally ordered with a gap $\leq 30$ frames and spatially compatible. We minimize the following merge cost:
\begin{equation}
    C_{merge} = \Delta t_{gap} + \frac{\|c_{last}^i - c_{first}^j\|_2}{\text{IoU}(box_{last}^i, box_{first}^j)}
\end{equation}
where $c$ denotes the bounding box centroid.
The overall tracking, post-processing, and ranking pipeline is visualized in Figure~\ref{fig:track_pipeline}.

\paragraph{TrackRank scoring function.}
To select the top $m \in [6, 10]$ salient entities per video, we compute a composite \textit{TrackRank} score $S_i$ that aggregates metrics for each track $i$ (all computed per video and normalized by the maximum over tracks). Unlike simple duration-based ranking, we use the following weighted formulation:
\begin{equation}\label{eq: track-ranker}
    S_i = \frac{\sum_{k} w_k \cdot M_{i,k}}{\sum_{k} w_k}
\end{equation}
The specific components and their empirically tuned weights are:
\begin{itemize}[leftmargin=*,itemsep=2pt]
    \item \textbf{Duration ($w=2.0$) \& Size ($w=1.0$)}: Favors tracks with sustained presence and higher average bounding box area.
    \item \textbf{Associated Objects ($w=2.0$)}: Normalized count of distinct non-person object classes overlapping the person's box (IoU $>0.2$).
    \item \textbf{Center Distance ($w=2.4$) \& Motion ($w=1.0$)}: Euclidean distance between first and last centroids, favoring traversals over stationary behavior.
    \item \textbf{Appearance Exceptionality ($w=2.2$)}: We quantify rarity as the normalized L1 distance from the dataset's average appearance in feature space (HSV and LBP histograms).
    \item \textbf{Scene Coverage ($w=1.5$)}: Area of the Convex Hull enclosing the track's boxes.
    \item \textbf{Quality Metrics}: Aggregates \textit{Average Confidence} ($w=0.8$, mean detection score), \textit{Smoothness} ($w=0.7$, computed as $1$ minus normalized acceleration magnitude to penalize jitter), and \textit{Aspect-Ratio Stability} ($w=0.5$, defined as $1$ minus the standard deviation of width/height ratios to penalize shape fluctuations).
\end{itemize}

\paragraph{Hard Filter Cascade.}
Prior to ranking, we enforce a hard filter: we keep only the COCO~\cite{lin2014coco} ``person'' class, require length $\geq 20$ frames, average area $\geq 5,500$ pixels, and require the track center to fall within the central safe region (frame cropped by 10\% margins) in at least 5 frames.

\paragraph{Diversity Sampling Strategy.}
To ensure diversity among the selected tracks, we employ a round-robin selection across rankings generated from multiple perturbed weight configurations ($\gamma \sim U(0.5, 1.5)$). This prevents redundancy (e.g., selecting visually identical pedestrians) and ensures a broad coverage of high-quality entities, which are subsequently manually validated to exclude phantom detections or identity switches.

\paragraph{Track Selection as Noise Control.} Per video, we select the top 6--10 tracks according to the TrackRank composite score (Eq.~\ref{eq: track-ranker}). Post-hoc analysis of the selected tracks' average detection confidence shows that they consistently fall within the top ${\sim}20\%$ most confident tracks per video, confirming that the ranking procedure implicitly filters out low-confidence, noise-prone trajectories before they can propagate errors into the question generation pipeline. The TrackRank selection process thus doubles as a noise control mechanism: by funneling only high-confidence, well-resolved trajectories into downstream task programming, it substantially mitigates the risk of identity switches, fragmented detections, or phantom tracks propagating into the ground-truth labels.

\subsection{Decoupled Descriptor Generation}
\label{app:descriptors}

\begin{figure*}[!t] 
    \centering
    \includegraphics[width=\textwidth]{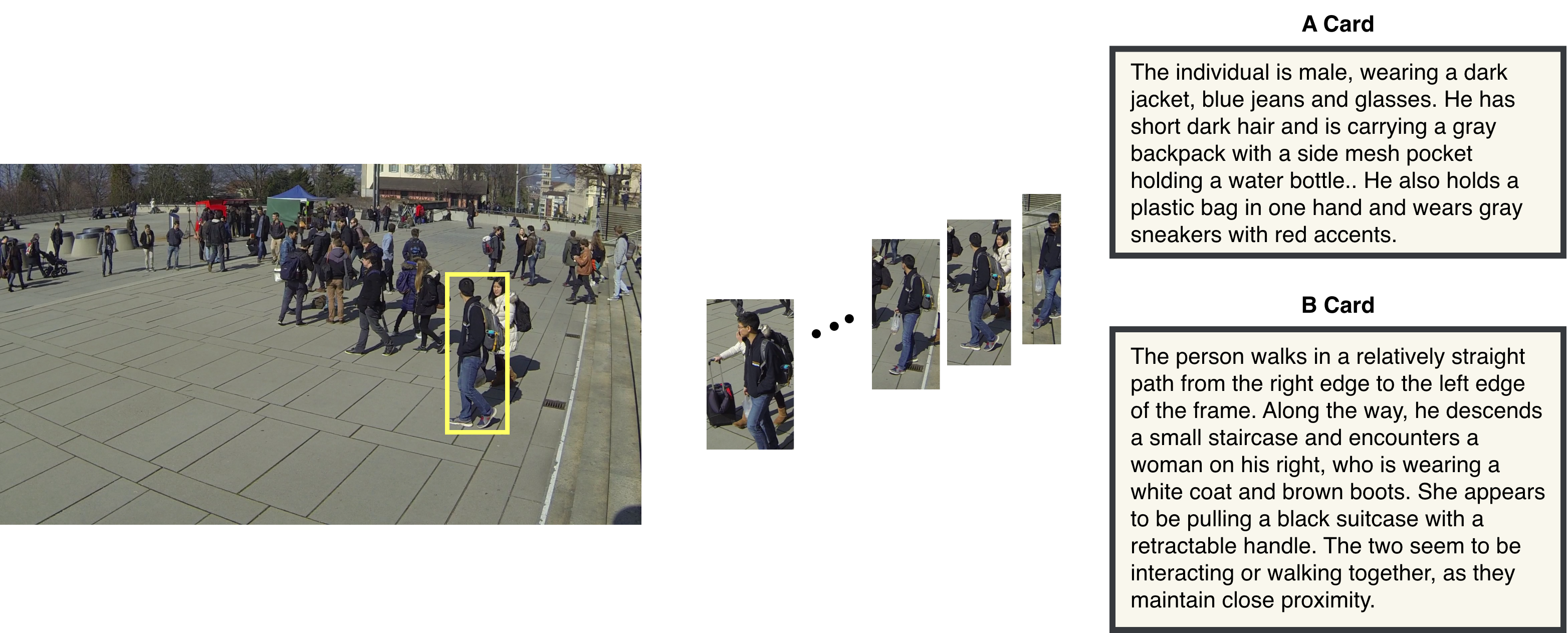} 
    \caption{\textbf{Example of disentangled A- and B-cards.} For a single tracked individual (highlighted trajectory in the top-left strip), we show the sampled frames and the corresponding appearance (A-card) and behavior (B-card) descriptions. The A-card captures only static visual attributes (clothing, colors, accessories, physique), while the B-card describes the person’s path, timing, and interactions over time without repeating appearance cues, enforcing the ``Look \& Separate'' principle.}
    \label{fig:ab_cards} 
\end{figure*}

\paragraph{A-card and B-card generation.}
For each selected track, we generate disentangled descriptions using GPT-4o~\cite{openai2024gpt4o}.
We sample 10-11 crops, reserving the first and last 20\% of the trajectory for Appearance (A-cards) and the middle 60\% for Behavior (B-cards). This ensures a temporal gap of at least 30 frames between appearance and behavior cues. An example of the resulting disentangled A- and B-cards for a single track is shown in Figure~\ref{fig:ab_cards}. We use the following prompt structure:

\medskip
\noindent\fbox{%
\parbox{\linewidth}{\small
\textbf{System prompt.} For the following tasks, use only your vision capabilities. When referring to directions, use the camera’s point of view.

\textbf{1. Person Description.} All images depict the same individual. In 2–4 sentences, describe their appearance in detail: clothing types and colors, accessories, hair, body build, and any distinctive features that make them easy to pick out. \emph{Do not mention position in the frame or any actions.}

\textbf{2. Path Description.} In 3–7 sentences, describe the person’s path and behavior over time. Mention the overall path shape, entry and exit edges, stops, and interactions. \emph{Do not repeat any appearance details from the first description.}
}%
}

\medskip

To visualize the output of this pipeline, Figure~\ref{fig:ab_cards} presents qualitative examples of the generated Appearance (A) and Behavior (B) cards alongside their corresponding tracked image crops. These examples highlights the effectiveness of the temporal split: the tracked visual crops from the start and end of the trajectory inform the static attribute descriptions in the A-card, while the central frames drive the dynamic action summaries in the B-card. This separation ensures that the descriptors remain disentangled.



\paragraph{Leakage prevention.}
To strictly enforce the ``Look \& Separate'' principle, we calculate the token-level Jaccard similarity between the generated A-card and B-card. 
We set the Jaccard threshold to 0.15 based on manual inspection: above this, descriptors often share explicit appearance/behavior leakage.

\subsection{Spatial Operations and Region Definitions}
\label{app:spatial-ops}

\paragraph{Entry/exit edge labeling.}
For tasks like \textit{Region-Localized People Counting (RLPC)}, we define entry and exit edges based on the position of a track’s centroid in its first and last frames. Let $c_t = (x_t, y_t)$ be the centroid at frame $t$ of a track with start frame $t_\text{start}$ and end frame $t_\text{end}$, and let $W,H$ denote the frame width and height. We say that a track enters through edge $e$ if $c_{t_\text{start}}$ lies in the corresponding edge band, and exits through edge $e'$ if $c_{t_\text{end}}$ lies in the band of $e'$. The top edge band is defined as $y < 0.3H$, the bottom as $y > 0.85H$, and the left/right edges as the outer $15\%$ of the width ($x < 0.15W$ and $x > 0.85W$, respectively).

\paragraph{Region-of-interest (ROI) membership.}
For \textit{[RLPC]}, we also define rectangular ROIs (e.g., frame halves or specific zones). A track is counted as visiting an ROI if, at any frame, at least 50\% of its bounding box area lies within the region (Intersection-Over-Box $\ge 0.5$). We count the unique track IDs that satisfy this predicate to derive people counts under spatial constraints. To absorb residual tracking noise (missed detections, fragmented tracks), multiple-choice options are reported as binned count ranges rather than exact integers. The bins are constructed so that the correct range spans approximately ±40\% around the true count on average, ensuring that minor tracking errors do not invalidate the ground truth while still requiring models to perform meaningful spatial counting, see Figure \ref{fig:rlpc_example} for example.

\paragraph{Duration computation (\textit{[MPDR]}).}
We compute visible-time intervals $(t_{start}, t_{end})$ for every track. Using interval algebra, we determine ground truth for questions such as ``Who stayed longest?'' or ``Who entered first?'' by comparing duration scalars ($t_{end} - t_{start}$) and timestamps.

\subsection{Scene Card Perturbations}
\label{app:perturbations}

\paragraph{Shot Segmentation and Description.}
We use \textit{TransNetV2}~\cite{soucek2020transnetv2} for shot boundary detection. To calibrate its reliability on our video corpus, we manually reviewed shot boundaries on 30 videos used for \textit{[TSO]} and \textit{[SVA]} tasks (34\% of videos contributing to these tasks). Comparing TransNetV2~\cite{soucek2020transnetv2} predictions against manual annotations yields F1 = 0.97, confirming that shot segmentation noise is negligible. For the \textit{[SVA]} task, faithful scene cards are generated via an MLLM using the following prompt:

\noindent\fbox{%
    \parbox{\linewidth}{%
        \small
        \textit{``Describe concisely the scene in one sentence without reference to the `scene', refer (if relevant) to the entities, genders and appearance (type and colors of hair/clothing/accessories) of each entity, occurrence, actions, background, and location.''}
    }%
}

\begin{figure*}[!t] 
    \centering
    \includegraphics[width=0.7\linewidth]{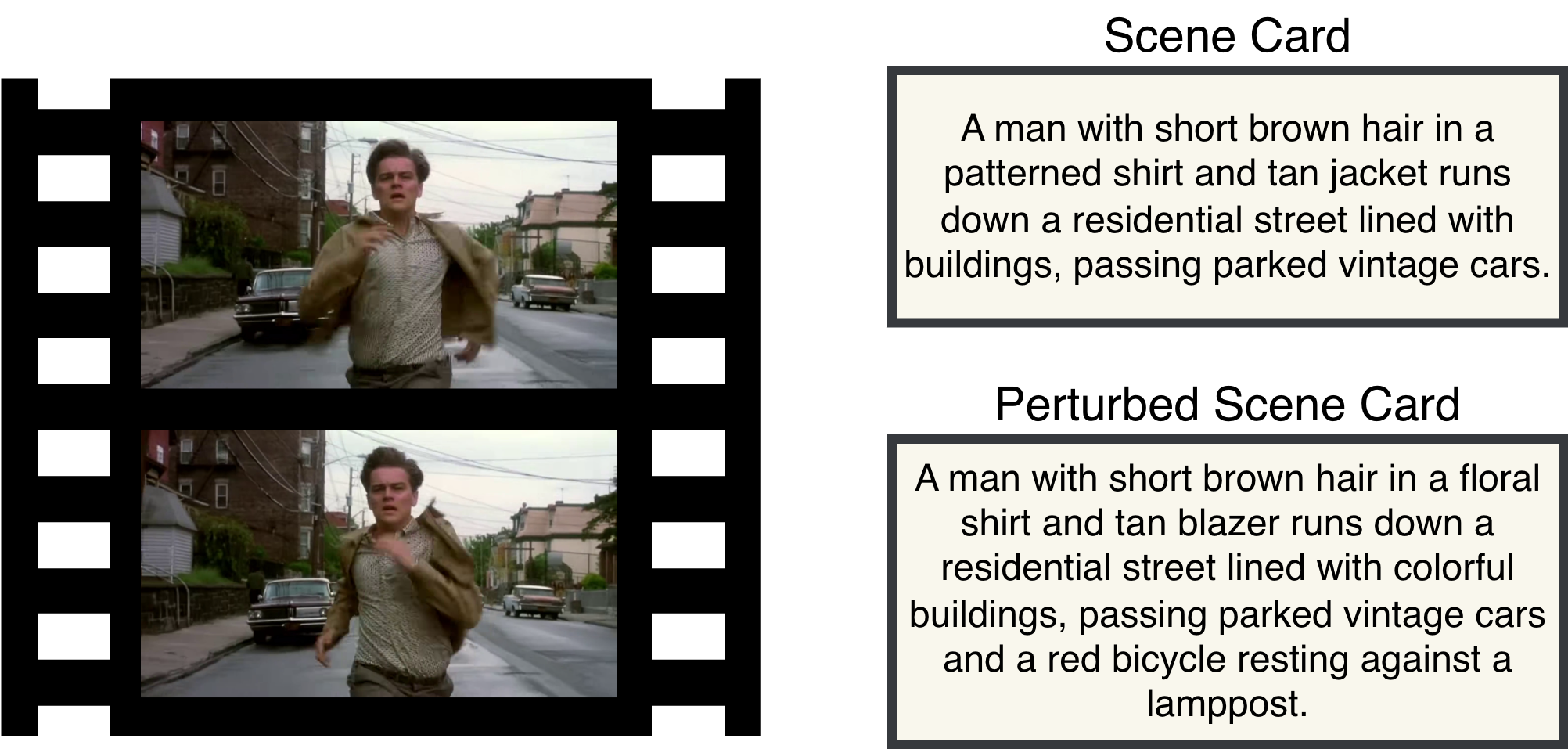} 
    \caption{\textbf{Faithful and perturbed scene cards for SVA.} The top card provides a faithful one-sentence description of a shot, mentioning the main actor, appearance, background, and motion. The bottom card is a perturbed variant where 2-5 atomic details (e.g., clothing pattern, background appearance, additional objects) are modified or added while remaining globally plausible. These pairs form positive and negative options in the Scene Verification \& Arrangement task, probing fine-grained scene-level sensitivity to small but visually significant details.}
    \label{fig:scene_cards} 
\end{figure*}


\paragraph{Perturbation Engine.}
To generate negative samples for \textit{[SVA]}, we prompt the model to modify faithful descriptions by altering 2-5 atomic details. The prompt constraints ensure:
\begin{itemize}
    \item \textbf{Modifications}: Change existing details (color, count, attributes).
    \item \textbf{Additions}: Insert plausible but absent elements (extra objects, background items).
    \item \textbf{Plausibility}: Changes must be false but highly plausible within the context of the video.
\end{itemize}

An example of a faithful scene card and its perturbed counterpart used for the \textit{[SVA]} task is shown in Figure~\ref{fig:scene_cards}.


\subsection{Corpus-Plausible Foil Generation}
\label{app:foils}

\paragraph{Ground Truth Integration.}
For tasks requiring verification of absence, we leverage human-verified event logs.
\begin{itemize}
    \item \textbf{False Action Memory (\textit{[FAM]})}: We sample a ``false'' action by pairing an object present in the video with an action from the corpus that does \textit{not} occur in the current video.
    \item \textbf{False Object Memory (\textit{[FOM]})}: We select an absent object from the corpus-wide index that is compatible with actions present in the video (e.g., if ``cutting'' occurs, ``carrot'' is a valid distractor if absent).
    \item \textbf{Action Counting (\textit{[AC]})}: Distractor counts are generated such that the correct count's rank varies uniformly across options.
    \item \textbf{Action Sequence Integrity (\textit{[ASII]})}: We sample a 5-event ground-truth timeline. Distractors are generated using two perturbation functions: \texttt{swap\_mid} (swapping two non-adjacent events) and \texttt{rotate} (shifting the sequence). Crucially, we verify against the event log that the perturbed timeline does not accidentally exist in the video.
\end{itemize}

\subsection{Text-Only Bias Filtering Details}
\label{app:text-filter}

\paragraph{Filtering procedure.}
To suppress language priors, we apply a rigorous Text-Only Filtering stage. We discard any question correctly answered by $\ge 3$ of 4 blind LLMs (Qwen2-7B~\cite{qwen2}, Qwen2.5-7B~\cite{qwen2_5}, Llama-3-8B~\cite{llama3}, and Vicuna-7B v1.5~\cite{vicuna15}). This step rejects approximately 10\% of candidates (e.g., questions answerable via object-color co-occurrence priors).

\subsection{Human Verification Protocol}
\label{app:human-verification}
Experts conduct verification on a stratified 15\% sample of instantiated questions to audit whether the construction pipeline preserves the intended evidential constraints. The audit focuses on three properties: (i) \textbf{minimum frame-set compliance}, i.e. confirming that the item requires at least three distinct frames; (ii) \textbf{uniqueness of the answer}, i.e. confirming the existence of a single objective ground-truth answer; and (iii) \textbf{descriptor disentanglement}, i.e. verifying that A-cards and B-cards do not leak information from one another. Items that violate any of these conditions are rejected. This process resulted in a 17.8\% rejection rate.

\subsection{Human Validation and Oracle-Frame Study}
\begin{table}[t]
\centering
\small
\setlength{\tabcolsep}{7pt}
\begin{tabular}{lcc}
\toprule
\textbf{Task} & \textbf{Full video (\%)} & \textbf{Oracle frames (\%)} \\
\midrule
\textit{[TSO]}  & 91.7 & 93.8 \\
\textit{[MPDR]} & 88.3 & 97.0 \\
\textit{[ASII]} & 86.7 & 95.8 \\
\textit{[AGBI]} & 95.8 & 98.1 \\
\textit{[AGAR]} & 95.0 & 98.3 \\
\textit{[AGLT]} & 92.5 & 97.4 \\
\textit{[FAM]}  & 84.2 & 92.8 \\
\textit{[SVA]}  & 84.2 & 94.8 \\
\textit{[FOM]}  & 87.5 & 96.3 \\
\textit{[MEGL]} & 85.8 & 95.4 \\
\textit{[AC]}   & 84.2 & 97.7 \\
\textit{[RLPC]} & 90.0 & 90.9 \\
\midrule
\textbf{Overall} & \textbf{88.8} & \textbf{95.7} \\
\bottomrule
\end{tabular}
\caption{\textbf{Per-task human accuracy.} Accuracy of human annotators in the full-video and oracle-frame settings across all HERBench tasks. In the full-video setting, annotators answer with unrestricted video access and free scrubbing. In the oracle-frame setting, annotators answer using only the curated oracle frame-set, without access to the source video.}
\label{tab:human_oracle_by_task}
\end{table}

To complement the construction-time audit above, we conducted two human studies that assess HERBench from two complementary perspectives: overall question answerability and oracle-evidence sufficiency. Each study used a separate group of 6 annotators.

\paragraph{Study design.}
In the \textbf{full-video} setting, annotators answered a shared set of 240 questions spanning all 12 tasks (20 questions per task) with unrestricted video access and free scrubbing. In the \textbf{oracle-frame} setting, annotators answered 2,160 questions using only the curated oracle frame-set provided by the benchmark construction pipeline, without access to the source video. Each oracle item was presented in the same format used by the oracle-based analysis in the main paper, namely the curated evidence frames together with distractor frames.

\paragraph{Results.}
Table~\ref{tab:human_oracle_by_task} reports the per-task accuracies. In the full-video setting, annotators achieved 88.8\% accuracy overall, with substantial inter-annotator agreement (Fleiss' $\kappa = 0.74$), indicating that the benchmark remains highly answerable for humans despite its high evidential demand. In the oracle-frame setting, annotators achieved 95.7\% accuracy overall, showing that the curated oracle frame-set is generally sufficient to resolve the question without access to the full temporal context. The largest improvements appear in tasks such as \textit{[AC]}, \textit{[SVA]}, and \textit{[MEGL]}, where the answer depends on sparse or temporally localized evidence.

\paragraph{Benchmark cleanup.}
After each oracle-frame response, annotators were shown the ground-truth answer and invited to flag problematic items. This process surfaced three types of issues: mis-indexed evidence frames, incorrect ground-truth labels, and genuinely ambiguous items. Specifically, annotators flagged 42/2160 items (1.9\%) for evidence mis-indexing, 18/2160 (0.8\%) for incorrect ground truth, and 58/2160 (2.7\%) as ambiguous. All flagged items were subsequently corrected or removed from the final release.

\subsection{Dataset Statistics}
\label{app:statistics}

\paragraph{Scale and Video Characteristics.}
HERBench comprises 26,806 questions derived from 336 unique videos. The videos feature substantial duration (avg. 395s, range 60-2100s) to ensure temporal dispersion of evidence. Sources include HD-EPIC~\cite{perrett2025hdepic}, WildTrack~\cite{wildtrack}, PersonPath22~\cite{personpath22}, and movie trailers.

\paragraph{Question Properties.}
The average question length is 65.5 tokens with a vocabulary of $\sim$7.3k unique word types. Questions are strictly balanced across 5 multiple-choice options. The mean temporal span of evidence required per question is 101.1 seconds.

\section{Extended Experimental Results \& Analysis}
\label{sec:extended_results}

We provide a deeper quantitative analysis of the challenges posed by HERBench, expanding on the MRFS metrics and frame selection ablation.

\subsection{Extended MRFS Analysis}
\label{sub:mrfs_analysis}

\paragraph{Per-Task MRFS.}

Table~\ref{tab:mrfs_per_task} details the Minimum Required Frame-Set statistics. We observe a distinct correlation between the reasoning scope of a task and its evidential requirement. Tasks requiring global chronology and the integration of multiple semantic units, specifically \textit{[TSO]} (Temporal Shot Ordering, MRFS 9.05), \textit{[FAM]} (False Action Memory, MRFS 6.77), and \textit{[SVA]} (Scene Verification, MRFS 6.74), naturally exhibit the highest MRFS. To answer these questions correctly, a model must aggregate evidence from widely dispersed video segments or perform an exhaustive search to verify absence, effectively precluding single-frame shortcuts. 

In contrast, tasks focused on local attributes or spatially constrained counting, such as \textit{[RLPC]} (Region-Localized People Counting, MRFS 3.11) and \textit{[AGAR]} (Attribute Recognition, MRFS 3.85), require fewer distinct frames. However, even these ``lower'' MRFS values demonstrate that reliance on a single frame is insufficient, confirming that HERBench successfully enforces multi-evidence integration even for localized tasks. The overall weighted mean MRFS of $5.49$ validates the benchmark's design goal: forcing models to look at multiple snapshots to derive correct answers.

\begin{table}[h]
\centering
\small
\setlength{\tabcolsep}{4.5pt}
\begin{tabular}{lrr}
\toprule
\textbf{Task} & \textbf{Total} & \textbf{Mean MRFS} \\
\midrule
\textit{[TSO]}  & 2123 & \textbf{9.05} \\
\textit{[MPDR]} & 2717 & 4.30 \\
\textit{[ASII]} & 2127 & 6.00 \\
\textit{[AGBI]} & 1226 & 3.81 \\
\textit{[AGAR]} & 876  & 3.85 \\
\textit{[AGLT]} & 2362 & 4.45 \\
\textit{[FAM]}  & 1962 & 6.77 \\
\textit{[SVA]}  & 4569 & 6.74 \\
\textit{[FOM]}  & 2022 & 5.14 \\
\textit{[MEGL]} & 3061 & 6.33 \\
\textit{[AC]}   & 1623 & 5.26 \\
\textit{[RLPC]} & 2138 & 3.11 \\
\midrule
\textbf{Total / Weighted Mean} & \textbf{26,806} & \textbf{5.49} \\
\bottomrule
\end{tabular}
\caption{\textbf{Per-task MRFS statistics} Computed with $x=16$ using Qwen2.5-VL~\cite{qwen25vl_72b} and AKS~\cite{tang2025aks}.}
\label{tab:mrfs_per_task}
\end{table}

\paragraph{MRFS vs Accuracy}
\begin{figure*}[h!] 
    \centering
    \includegraphics[width=\textwidth]{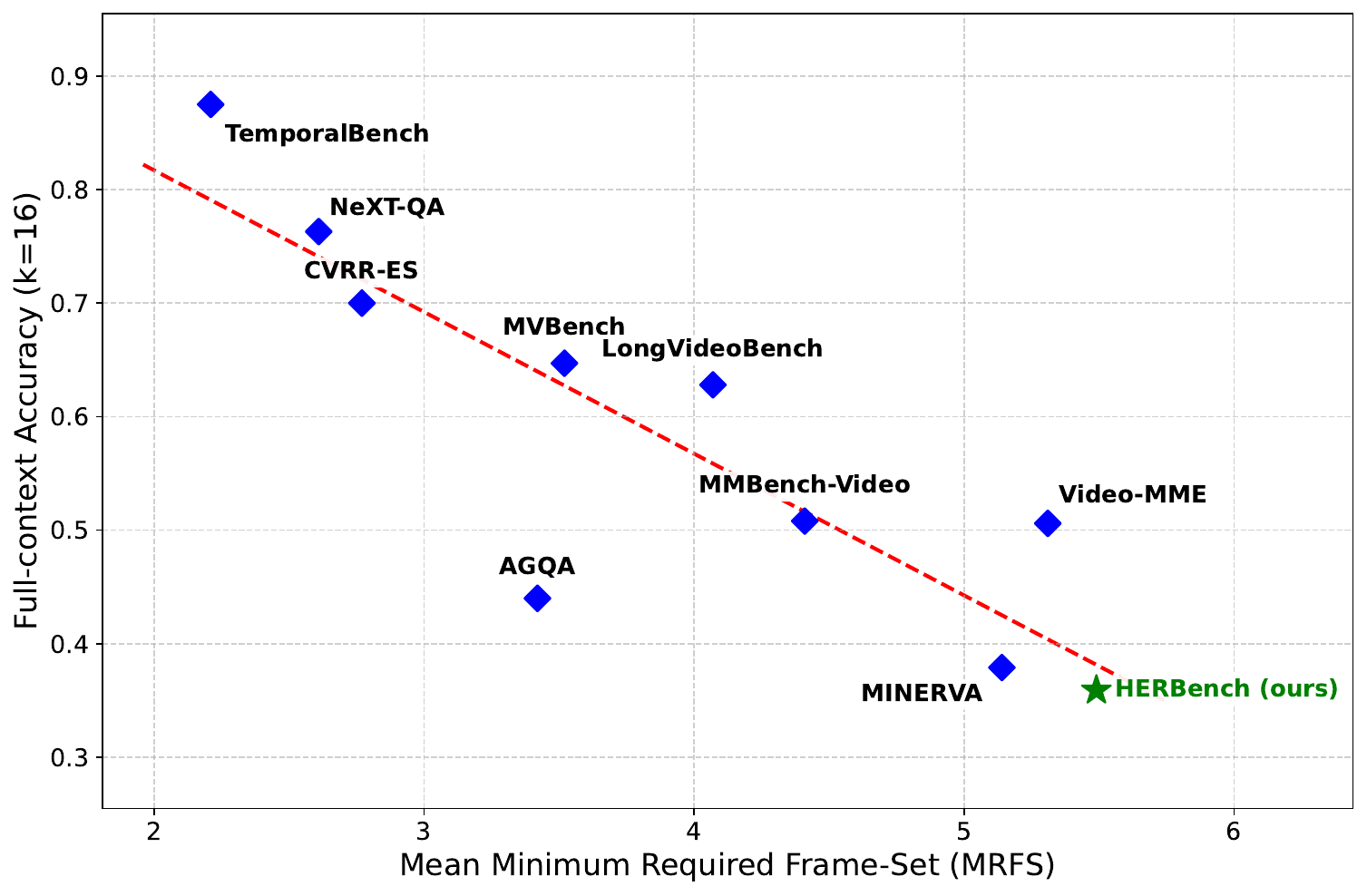}
    \caption{\textbf{Impact of Evidential Requirement on Model Accuracy.} We plot the Mean Minimum Required Frame-Set (MRFS) against Full-context Accuracy ($k=16$), measured using Qwen 2.5 VL 7B~\cite{qwen25vl_72b}, across ten video QA benchmarks. The dashed line indicates the fitted trend. A clear inverse correlation emerges: as the evidential burden increases (higher MRFS), model performance tends to decrease ($R^2 = 0.82$). HERBench (green star) occupies the high-demand end of the spectrum ($\text{MRFS}\ 5.49$), highlighting the challenges current Video-LLMs face in multi-evidence integration relative to lower-requirement benchmarks such as TemporalBench and NeXT-QA.}
    \label{fig:mrfs_accuracy} 
\end{figure*}

As illustrated in Figure~\ref{fig:mrfs_accuracy}, there is a pronounced inverse relationship between the evidential demand of a benchmark—quantified by the Mean MRFS—and the performance of state-of-the-art Video-LLMs. Benchmarks with low evidential requirements, such as TemporalBench ($2.21\ \text{MRFS}$, $87.5\%$) and NeXT-QA~\cite{xiao2021nextqa} ($2.61\ \text{MRFS}$, $76.3\%$), allow Qwen 2.5 VL 7B~\cite{qwen25vl_72b} to achieve relatively high accuracy, possibly due to the feasibility of single-frame shortcuts or language priors. Mid-range benchmarks—CVRR-ES, MVBench, LongVideoBench, and MMBench-Video—cluster between $2.8$--$4.4\ \text{MRFS}$ with accuracies that progressively decline from $70.0\%$ to $50.8\%$, tracing the fitted trend closely. At the high-demand end, MINERVA ($5.14\ \text{MRFS}$, $37.9\%$), Video-MME ($5.31\ \text{MRFS}$, $50.6\%$), and HERBench ($5.49\ \text{MRFS}$, $35.9\%$) impose substantially greater evidential burdens, coinciding with markedly lower accuracies. Notably, AGQA ($3.42\ \text{MRFS}$, $44.0\%$) falls below the trend line, suggesting that factors beyond evidential density—such as compositional question complexity—can independently depress performance. The broadened comparison across ten benchmarks strengthens the evidence for a \textit{fusion deficit} in current architectures: while models may be effective at retrieving isolated frames, their capacity for compositional reasoning degrades consistently as the number of required evidence pieces grows. HERBench, positioned at the extreme of this spectrum, is specifically designed to stress-test this bottleneck by requiring the integration of non-redundant, temporally dispersed cues.

\begin{table*}[t!]
\centering
\scriptsize
\caption{\textbf{Frame Selection Ablation.} Accuracy (\%) on a random subsample of questions using InternVL3.5~\cite{internvl35_8b_2025}, Qwen3-VL~\cite{bai2025qwen3vl}, and Ovis-2.5~\cite{ovis25_9b_2025} with Uniform, Vanilla-BLIP~\cite{li2022blip}, BOLT-ITS~\cite{liu2025bolt}, AKS~\cite{tang2025aks}, and GT Frames (OF) selectors. GT Frames represents the upper bound with manually curated evidence.}
\label{tab:frame_ablation_full}
\setlength{\tabcolsep}{2.5pt}
\resizebox{\textwidth}{!}{%
\begin{tabular}{l|l|cccccccccccc|c}
\toprule
\textbf{Model} & \textbf{Selector} & \textbf{AC} & \textbf{AGAR} & \textbf{AGBI} & \textbf{AGLT} & \textbf{ASII} & \textbf{FAM} & \textbf{FOM} & \textbf{MEGL} & \textbf{MPDR} & \textbf{RLPC} & \textbf{SVA} & \textbf{TSO} & \textbf{Mean} \\
\midrule
\multirow{5}{*}{\textbf{InternVL3.5}} 
 & Uniform & 23.0 & 75.0 & 77.0 & 70.0 & 32.0 & 30.0 & 30.0 & 34.0 & 48.0 & 27.0 & 23.0 & 41.0 & 42.7 \\
 & Vanilla-BLIP & 26.0 & 74.0 & 76.0 & 71.0 & 27.0 & 27.0 & 29.0 & 28.0 & 46.0 & 33.0 & 41.0 & 43.0 & 42.1 \\
 & BOLT-ITS & 22.0 & 72.0 & 74.0 & 71.0 & 20.0 & 27.0 & 33.0 & 30.0 & 48.0 & 33.0 & 27.0 & 36.0 & 41.1 \\
 & AKS & 27.0 & 66.0 & 77.0 & 74.0 & 36.0 & 29.0 & 30.0 & 35.0 & 54.0 & 33.0 & 33.0 & 17.0 & 42.7 \\
 & \textbf{GT Frames} & 24.0 & 81.0 & 81.0 & 79.0 & 20.0 & 50.0 & 39.0 & 27.0 & 52.0 & 32.0 & 37.0 & 53.0 & \textbf{47.8} \\
\midrule
\multirow{5}{*}{\textbf{Qwen3-VL}} 
 & Uniform & 26.0 & 67.0 & 78.0 & 68.0 & 34.0 & 30.0 & 24.0 & 16.0 & 36.0 & 23.0 & 50.0 & 0.0 & 37.7 \\
 & Vanilla-BLIP & 27.0 & 71.0 & 76.0 & 66.0 & 26.0 & 24.0 & 23.0 & 30.0 & 37.0 & 19.0 & 56.0 & 0.0 & 37.9 \\
 & BOLT-ITS & 25.0 & 68.0 & 75.0 & 66.0 & 27.0 & 21.0 & 33.0 & 30.0 & 38.0 & 21.0 & 57.0 & 0.0 & 38.4 \\
 & AKS & 24.0 & 65.0 & 73.0 & 69.0 & 29.0 & 22.0 & 27.0 & 20.0 & 35.0 & 22.0 & 49.0 & 0.0 & 36.2 \\
 & \textbf{GT Frames} & 24.0 & 69.0 & 73.0 & 71.0 & 35.0 & 50.0 & 25.0 & 24.0 & 36.0 & 21.0 & 61.0 & 3.0 & \textbf{41.0} \\
\midrule
\multirow{5}{*}{\textbf{Ovis-2.5}} 
 & Uniform & 25.0 & 79.0 & 81.0 & 71.0 & 34.0 & 35.0 & 34.0 & 35.0 & 38.0 & 21.0 & 65.0 & 0.0 & 43.1 \\
 & Vanilla-BLIP & 32.0 & 77.0 & 83.0 & 69.0 & 17.0 & 29.0 & 33.0 & 37.0 & 44.0 & 19.0 & 58.0 & 0.0 & 41.6 \\
 & BOLT-ITS & 35.0 & 78.0 & 82.0 & 70.0 & 17.0 & 28.0 & 33.0 & 38.0 & 46.0 & 18.0 & 60.0 & 0.0 & 42.1 \\
 & AKS & 25.0 & 76.0 & 80.0 & 74.0 & 31.0 & 39.0 & 39.0 & 30.0 & 49.0 & 17.0 & 51.0 & 0.0 & 42.6 \\
 & \textbf{GT Frames} & 30.0 & 85.0 & 84.0 & 80.0 & 23.0 & 60.0 & 39.0 & 40.0 & 41.0 & 21.0 & 68.0 & 4.0 & \textbf{47.9} \\
\bottomrule
\end{tabular}
}%
\end{table*}

\vspace{2em}

\subsection{Full Frame-Selection Ablation}
\label{sub:frame_ablation}


To more precisely disentangle the role of evidence retrieval from that of
multi-evidence fusion, we perform an extensive ablation over five frame
selection strategies—Uniform, Vanilla-BLIP~\cite{li2022blip}, BOLT-ITS~\cite{liu2025bolt}, AKS~\cite{tang2025aks}, and
Oracle Frames (OF)—and evaluate their effect across all twelve HERBench
tasks (Table~\ref{tab:frame_ablation_full}). 
Overall, learned strategies such as BOLT-ITS and AKS provide moderate gains
over Uniform sampling, reflecting their ability to prioritize query-relevant
frames while maintaining broader temporal coverage. However, their improvements
are uneven across tasks: both methods show the largest benefits in
sparse-evidence settings such as \textit{[TSO]} and \textit{[FAM]}, where the
critical evidence may appear only briefly within long videos. The
oracle-based setting establishes an upper bound by supplying the manually
curated evidence frames used during dataset construction. As shown in the
rightmost column of Table~\ref{tab:frame_ablation_full}, all three representative models experience
non-trivial but still limited performance improvements in the OF regime
(typically $+3$-$6$ absolute accuracy points relative to the best learned
selector). 

Importantly, the OF results highlight two key phenomena. First, even perfect 
access to the relevant frames does not resolve the majority of model failures:
fusion-bound tasks such as \textit{[AC]}, \textit{[RLPC]}, and \textit{[MEGL]}
remain bottlenecks with accuracies barely above chance, indicating that
retrieval is not the sole limiting factor. Second, improvements under OF are
disproportionately large for temporally global tasks such as \textit{[TSO]} and
\textit{[SVA]}, where correct reasoning requires coordinating multiple distant,
non-overlapping visual clues. Here retrieval quality is a dominant factor, and
learned selectors struggle to consistently surface all required frames.
However, the inability of models to capitalize fully on oracle-quality evidence
emphasizes that multi-frame integration itself remains a major unresolved
challenge. Taken together, these results reinforce a two-stage deficit: (i) an
\emph{evidence retrieval bottleneck}, where existing selectors fail to reliably
surface all critical cues, and (ii) a more fundamental \emph{fusion bottleneck},
where models fail to combine available cues even when retrieval uncertainty is
eliminated. HERBench’s high evidential density and stringent cue separation
make both deficits sharply visible, underscoring the need for future
MLLMs to improve not only frame selection but also the downstream
mechanisms for multi-cue aggregation.

\subsection{MRFS robustness across backbones and frame selectors}
\label{app:mrfs_robustness}

\begin{table*}[t]
    \centering
    \caption{MRFS robustness across models and keyframe selectors. Benchmarks: NExT-QA~\cite{xiao2021nextqa}, MVBench~\cite{li2024mvbench}, LongVideoBench~\cite{longvideobench2024}, and HERBench.}
    \label{tab:mrfs_robustness}
    
    \small
    \setlength{\tabcolsep}{4.5pt}
    \renewcommand{\arraystretch}{1.1}
    \begin{tabular}{lccccccc}
    \toprule
    \textbf{Benchmark} &
    \multicolumn{4}{c}{\textbf{Model on AKS (MRFS / Acc.)}} &
    \multicolumn{3}{c}{\textbf{Selector on Qwen2.5-VL (MRFS / Acc.)}} \\
    \cmidrule(lr){2-5}\cmidrule(lr){6-8}
    & \textbf{Qwen2.5-VL} & \textbf{Gemini2.5F} & \textbf{LLaVA-OV-1.5} & \textbf{GPT-4o}
    & \textbf{AKS} & \textbf{BOLT} & \textbf{$T^*$} \\
    \midrule
    NExT-QA        & 2.61 / 65.79 & 3.81 / 76.29 & 3.45 / 68.77 & 3.85 / 70.81 & 2.61 / 65.79 & 3.64 / 75.33 & 2.78 / 64.56 \\
    MVBench        & 3.52 / 56.71 & 3.92 / 57.41 & 3.69 / 54.24 & 4.09 / 54.42 & 3.52 / 56.71 & 3.73 / 56.28 & 3.53 / 55.53 \\
    LongVideoBench & 4.07 / 41.38 & 4.57 / 64.59 & 4.50 / 61.49 & 4.92 / 59.01 & 4.07 / 41.38 & 4.89 / 49.14 & 4.11 / 48.59 \\
    \rowcolor{red!10}
    \textbf{HERBench} & \textbf{5.49 / 35.91} & \textbf{5.68 / 38.74} & \textbf{5.67 / 36.50} & \textbf{5.81 / 38.65} & \textbf{5.49 / 35.91} & \textbf{5.72 / 42.08} & \textbf{5.43 / 43.41} \\
    \bottomrule
    \vspace{1em}
    \end{tabular}
    \end{table*}

To assess whether MRFS-based benchmark comparison is sensitive to the choice of backbone or selector, we evaluate MRFS under a range of configurations: four backbone models (Qwen2.5-VL~\cite{qwen25vl_72b}, Gemini 2.5 Flash~\cite{comanici2025gemini}, LLaVA-OV-1.5~\cite{llava_onevision15_8b_2025}, and GPT-4o~\cite{openai2024gpt4o}) with AKS~\cite{tang2025aks}, and three selectors (AKS, BOLT~\cite{liu2025bolt}, and $T^\star$~\cite{ye2025re}) with Qwen2.5-VL (see Tab.~\ref{tab:mrfs_robustness}). Here, $T^\star$ serves as a non-CLIP baseline. All results are computed on 50\% stratified random samples from each benchmark. Across all tested settings, the benchmark ordering remains stable, with HERBench consistently yielding the highest MRFS, followed by LongVideoBench, MVBench, and NExT-QA. These results support MRFS as a robust measure of dataset-level evidential requirement.

\vspace{3em}
\section{Illustrative Examples for All Tasks}

This section provides qualitative examples for all twelve HERBench tasks, each figure displays \emph{one representative structured question} for the corresponding task. However, each task in HERBench contains \emph{many distinct question structures and evidential templates}, and the examples below illustrate only a single instance of the broader variability present in the dataset.

\begin{figure*}[t!] 
    \centering
    \includegraphics[width=\textwidth]{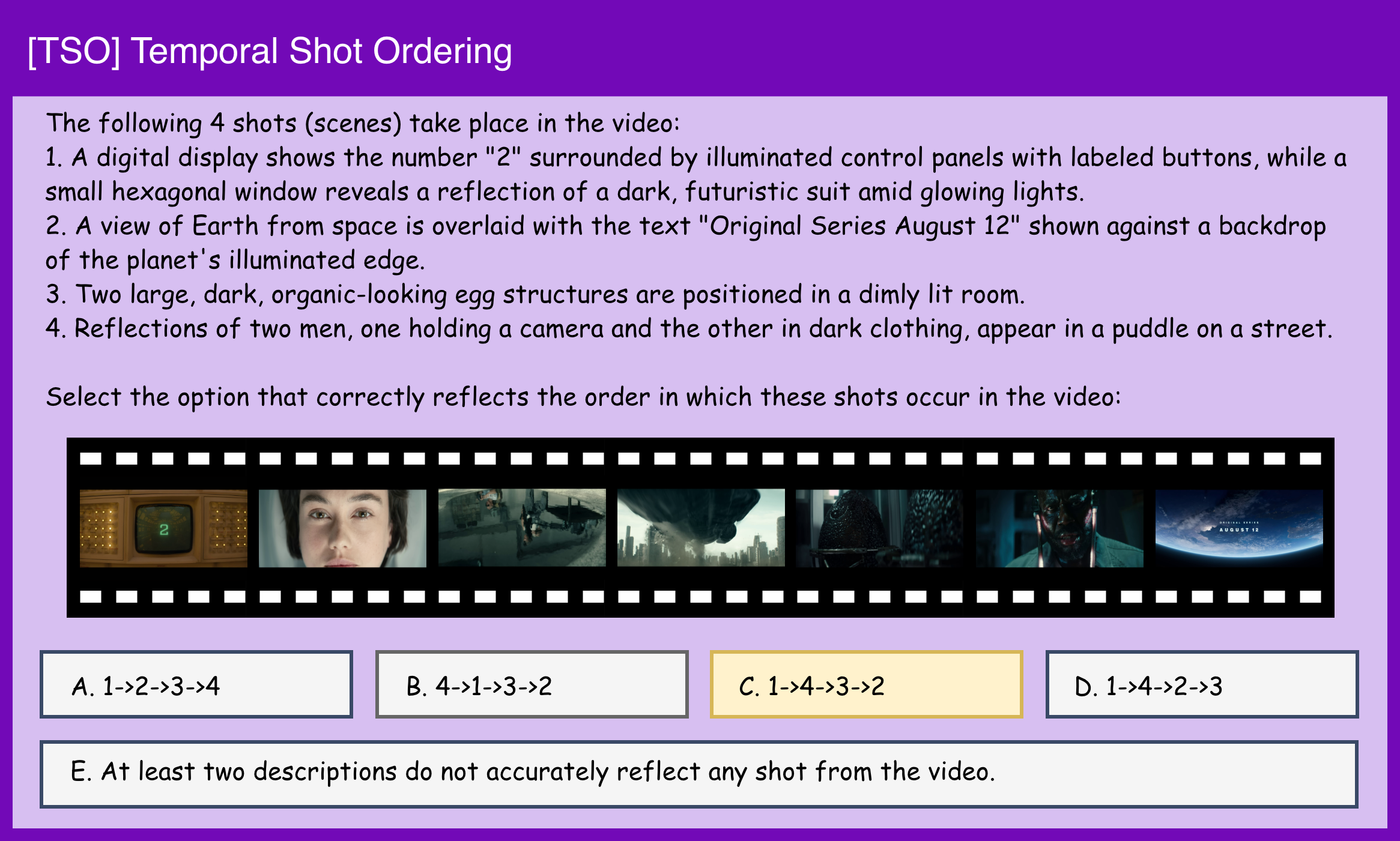}
    \caption{}
    \label{fig:tso_example} 
\end{figure*}

\begin{figure*}[!t]
    \centering
    \includegraphics[width=\textwidth]{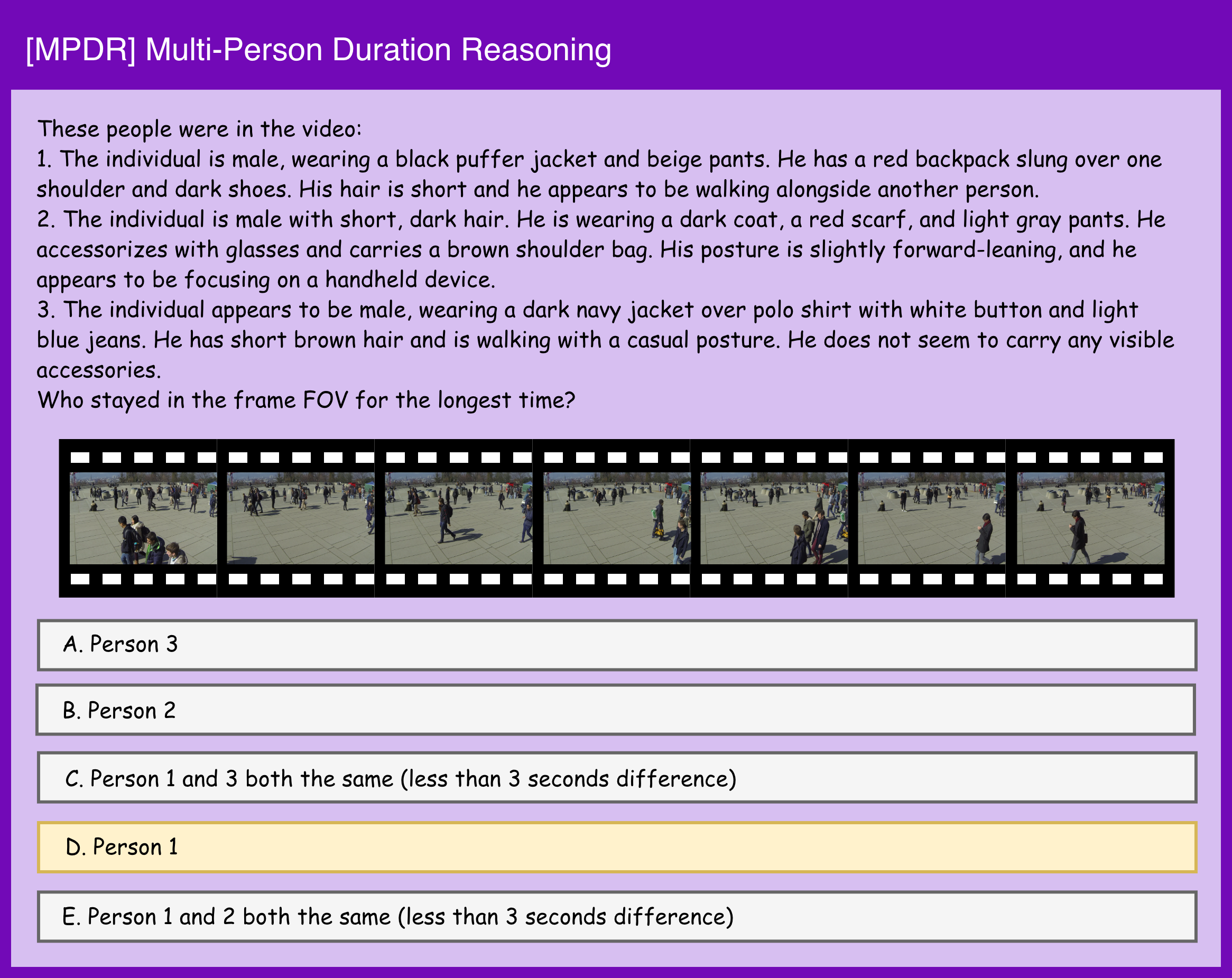}
    \caption{}
    \label{fig:mpdr_example} 
\end{figure*}

\begin{figure*}[!t]
    \centering
    \includegraphics[width=\textwidth]{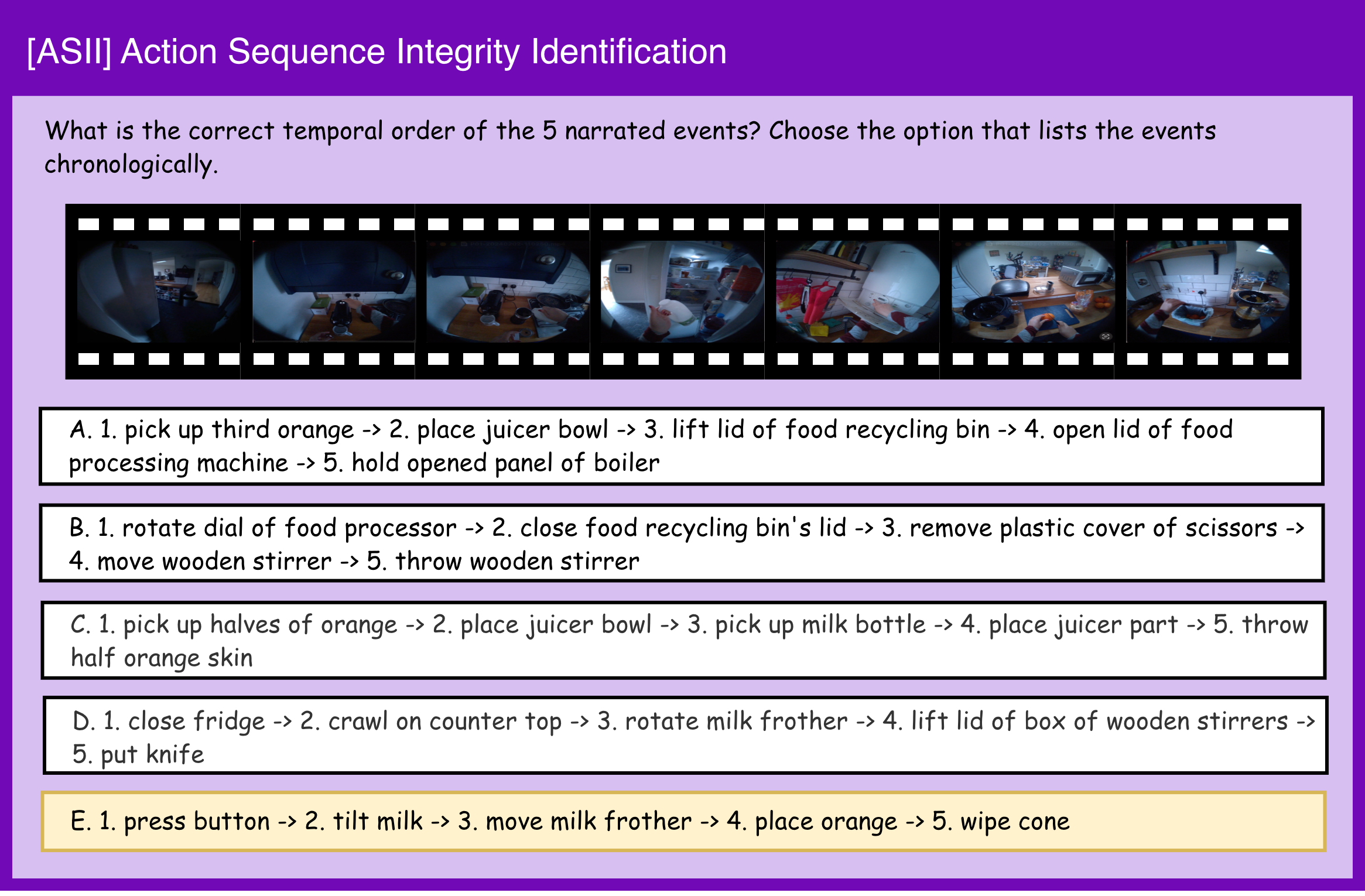}
    \caption{}
    \label{fig:asii_example} 
\end{figure*}

\paragraph{Temporal Reasoning \& Chronology.}
Figure~\ref{fig:tso_example} presents an example of the \textit{Temporal Shot Ordering ([TSO])} task, which requires reconstructing the chronological order of four non-overlapping shots. 
Figure~\ref{fig:mpdr_example} shows the \textit{Multi-Person Duration Reasoning ([MPDR])} task, where models must compare visible-time intervals across multiple individuals. 
Figure~\ref{fig:asii_example} illustrates the \textit{Action Sequence Integrity \& Identification ([ASII])} task, requiring identification of the correct sequence among plausible permutations of narrated events.

\begin{figure*}[!t]
    \centering
    \includegraphics[width=\textwidth]{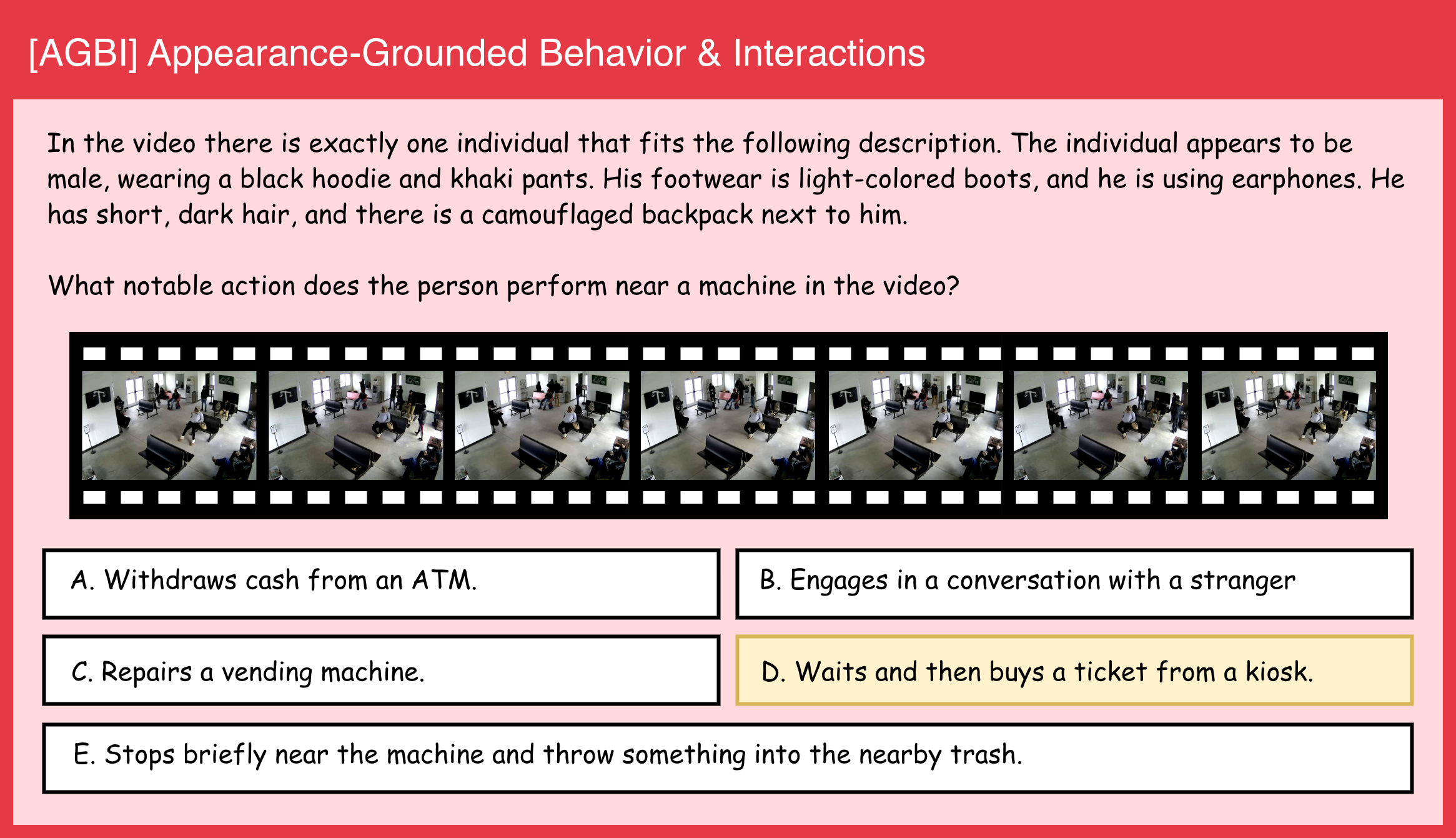}
    \caption{}
    \label{fig:agbi_example} 
\end{figure*}

\begin{figure*}[!t]
    \centering
    \includegraphics[width=\textwidth]{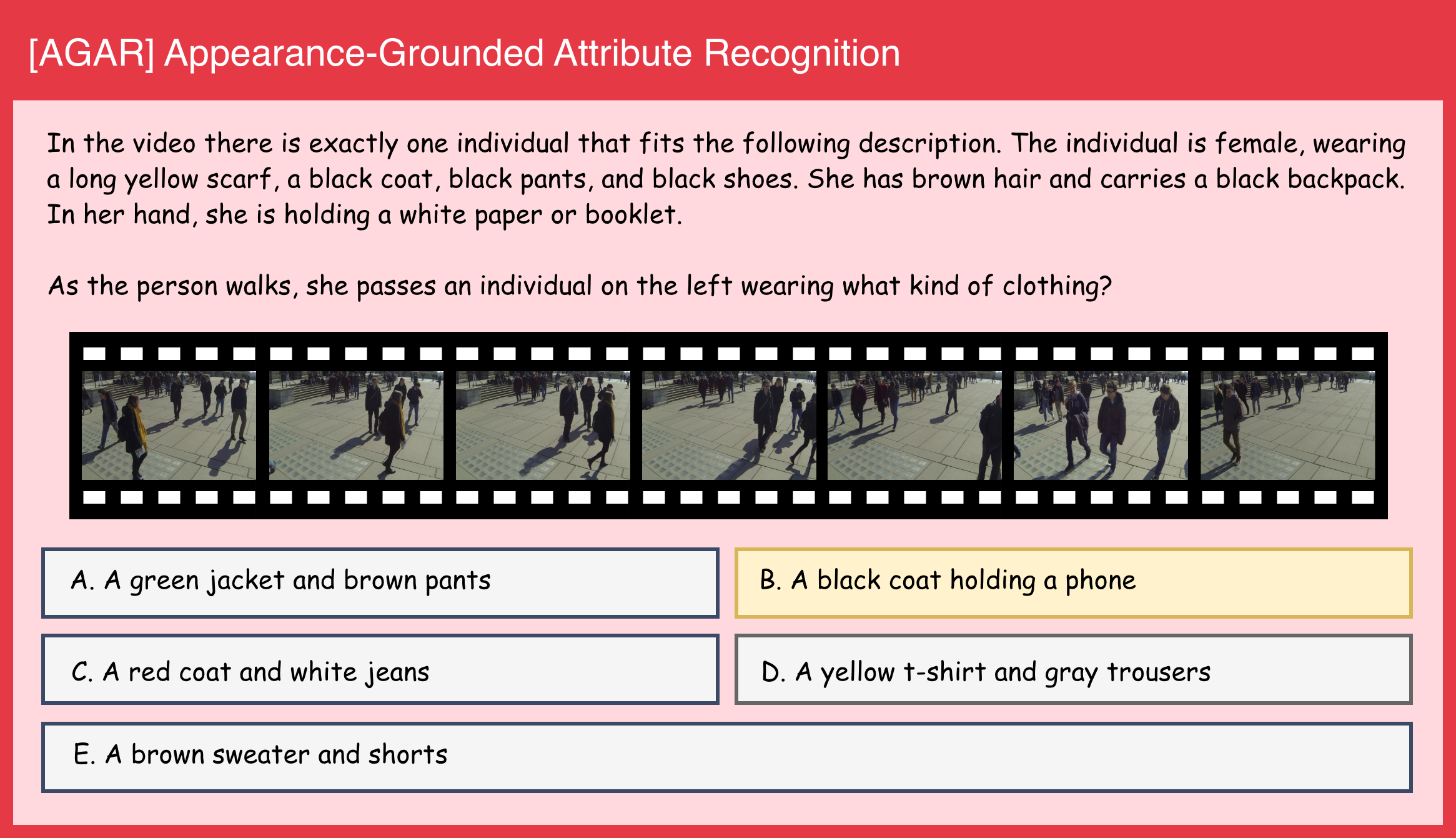}
    \caption{}
    \label{fig:agar_example} 
\end{figure*}

\begin{figure*}[!t]
    \centering
    \includegraphics[width=\textwidth]{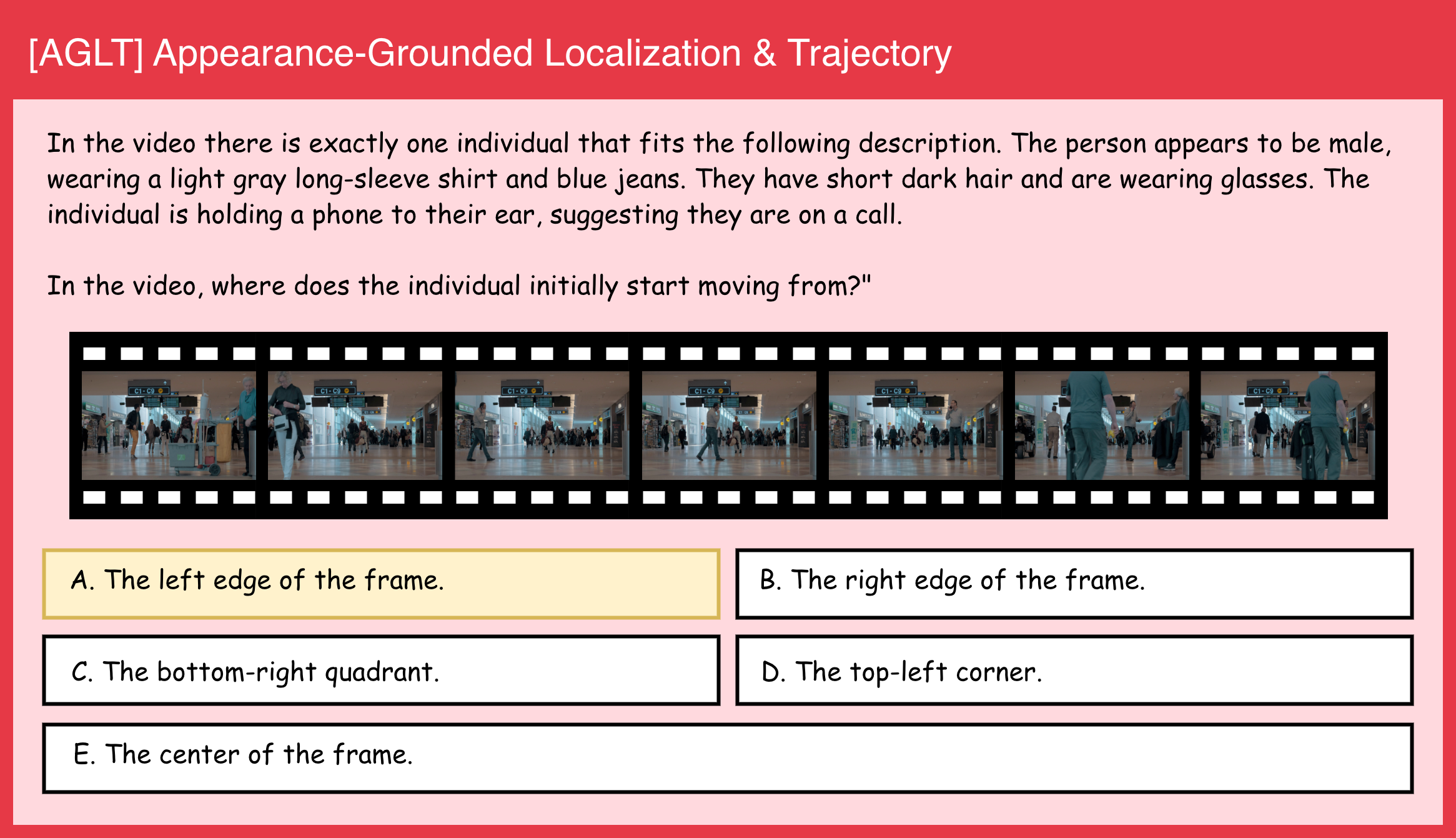}
    \caption{}
    \label{fig:aglt_example} 
\end{figure*}

\paragraph{Referring \& Tracking.}
Figure~\ref{fig:agbi_example} shows the \textit{Appearance-Grounded Behavior Interactions ([AGBI])} task, where models must track a target described only by appearance and determine who interacts with them. 
Figure~\ref{fig:agar_example} provides an example of the \textit{Appearance-Grounded Attribute Recognition ([AGAR])} task, requiring attribute extraction anchored to the tracked target. 
Figure~\ref{fig:aglt_example} illustrates the \textit{Appearance-Grounded Localization Trajectory ([AGLT])} task, where the model must infer how the target enters or exits the scene.

\begin{figure*}[!t]
    \centering
    \includegraphics[width=\textwidth]{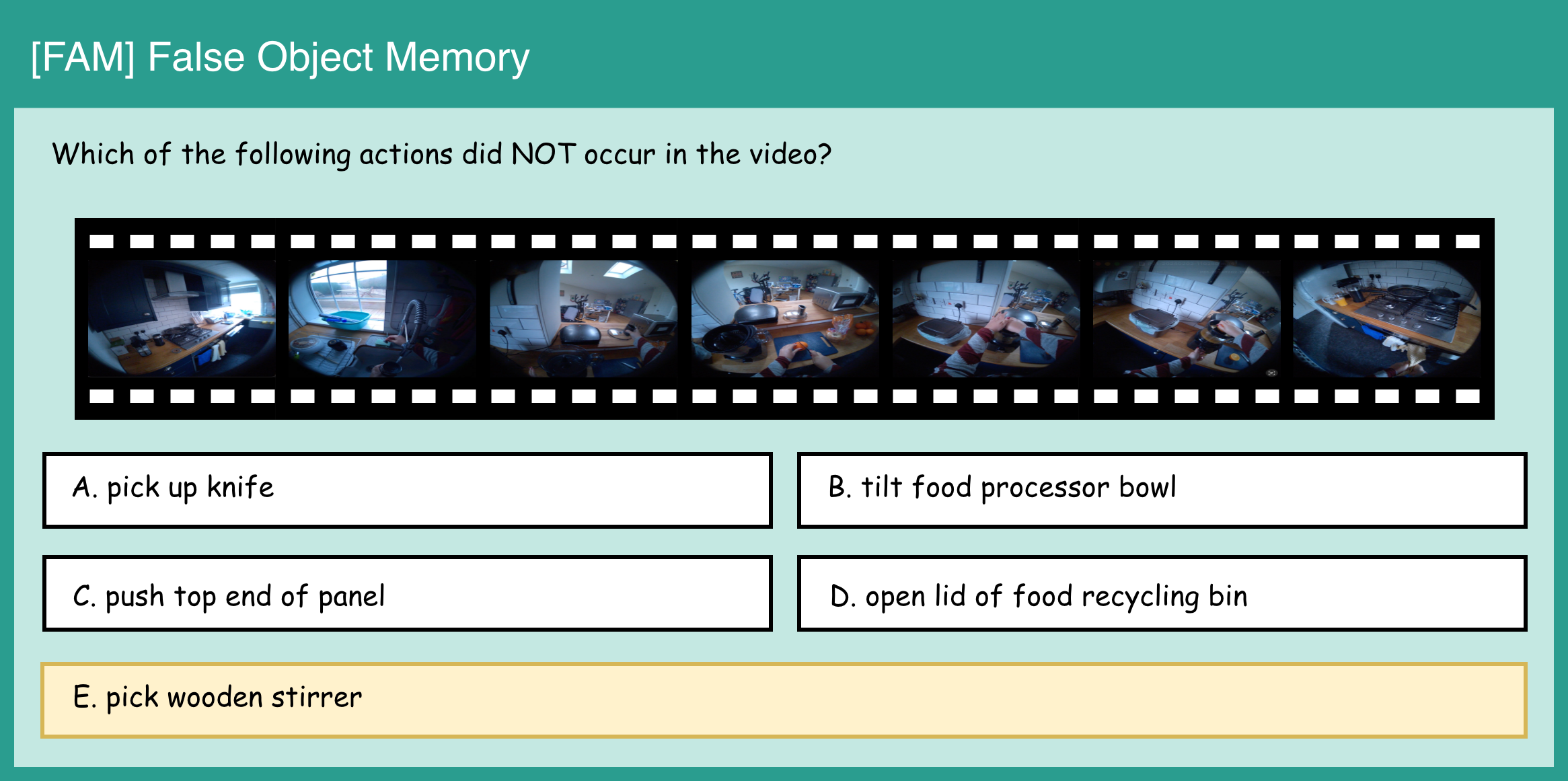}
    \caption{}
    \label{fig:fam_example} 
\end{figure*}

\begin{figure*}[!t]
    \centering
    \includegraphics[width=\textwidth]{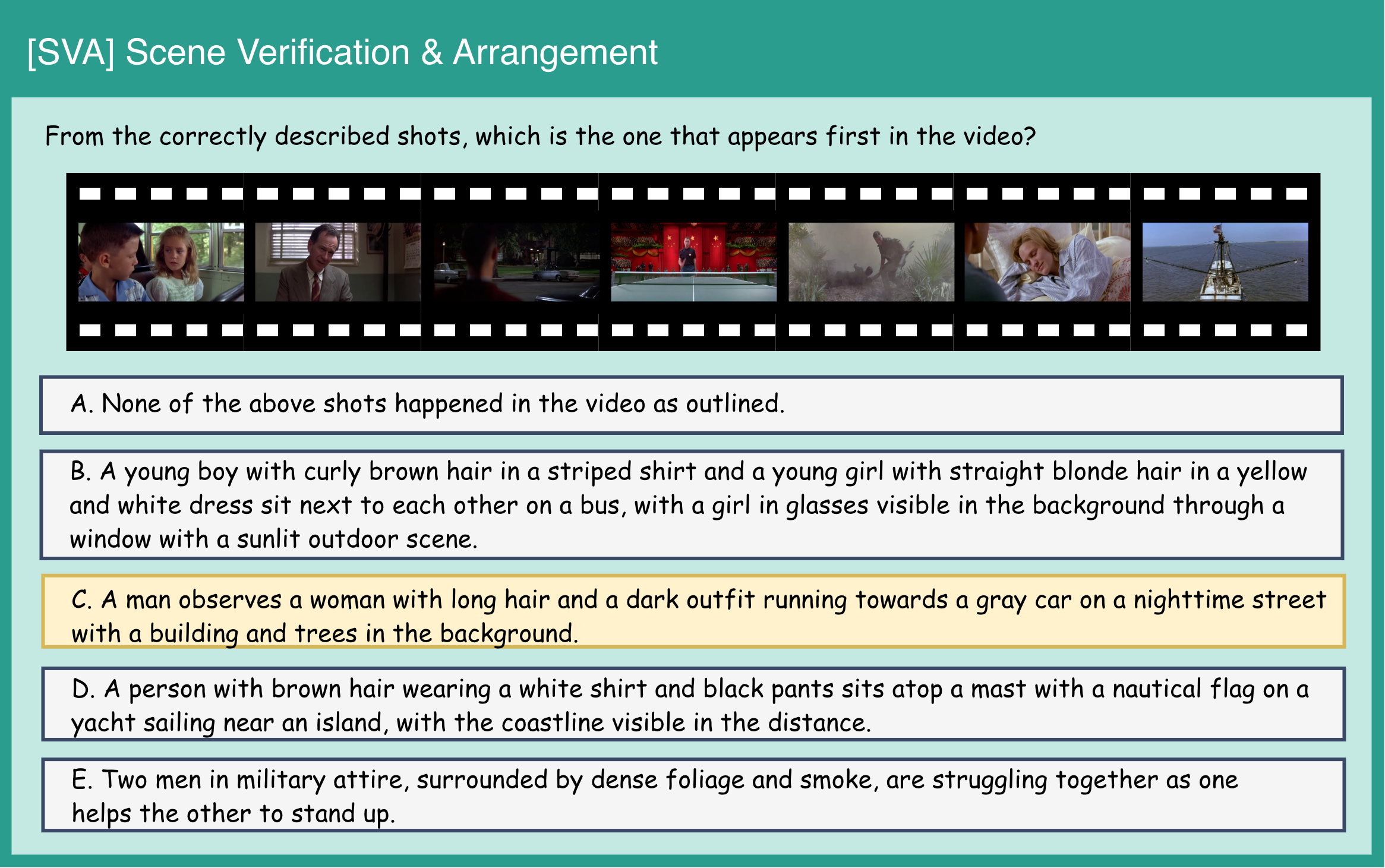}
    \caption{}
    \label{fig:sva_example} 
\end{figure*}

\begin{figure*}[!t]
    \centering
    \includegraphics[width=\textwidth]{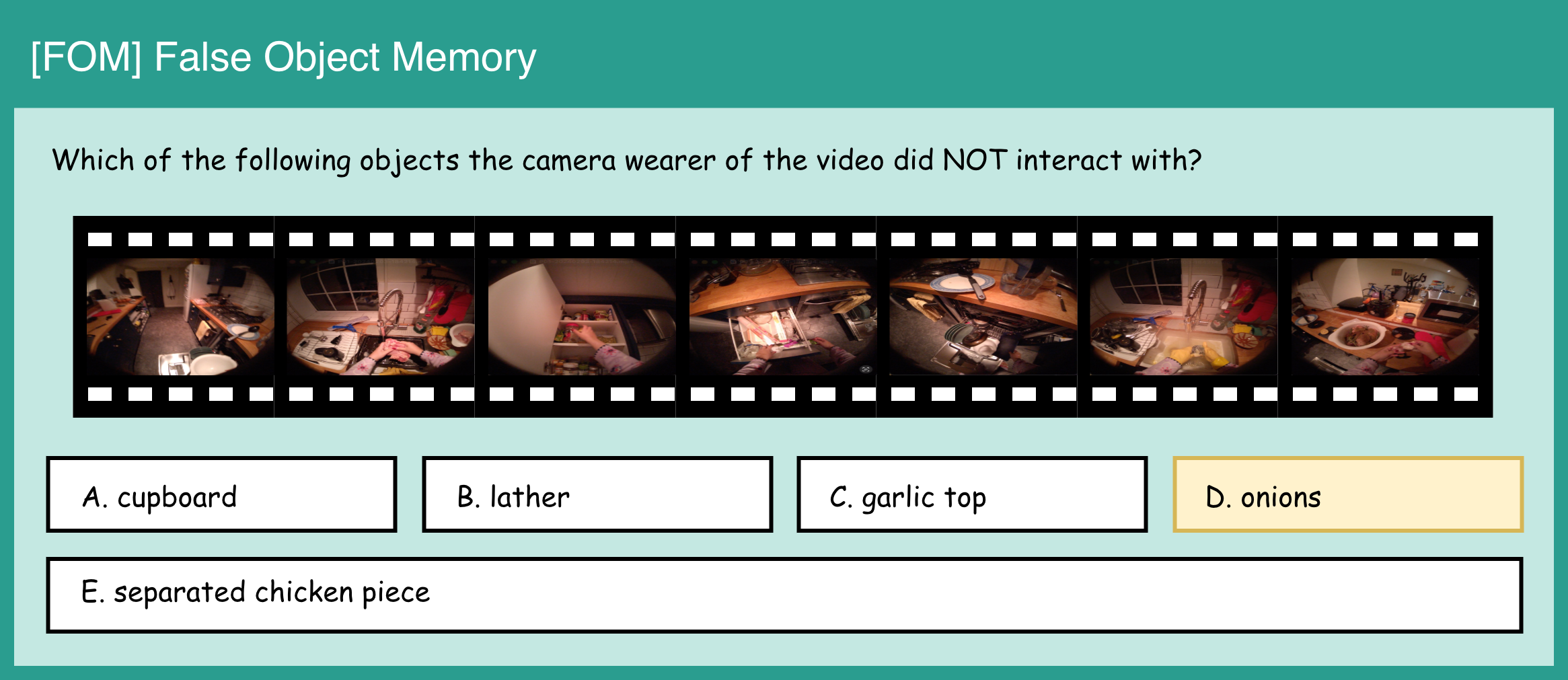}
    \caption{}
    \label{fig:fom_example} 
\end{figure*}

\paragraph{Global Consistency \& Verification.}
Figure~\ref{fig:fam_example} presents the \textit{False Action Memory ([FAM])} task, requiring verification of which plausible action did \emph{not} occur in the video. 
Figure~\ref{fig:sva_example} shows the \textit{Scene Verification \& Arrangement ([SVA])} task, combining faithful and perturbed shot descriptions to assess fine-grained scene-level verification and ordering. 
Figure~\ref{fig:fom_example} depicts the \textit{False Object Memory ([FOM])} task, requiring identification of a plausible but absent object interaction.

\begin{figure*}[!t]
    \centering
    \includegraphics[width=\textwidth]{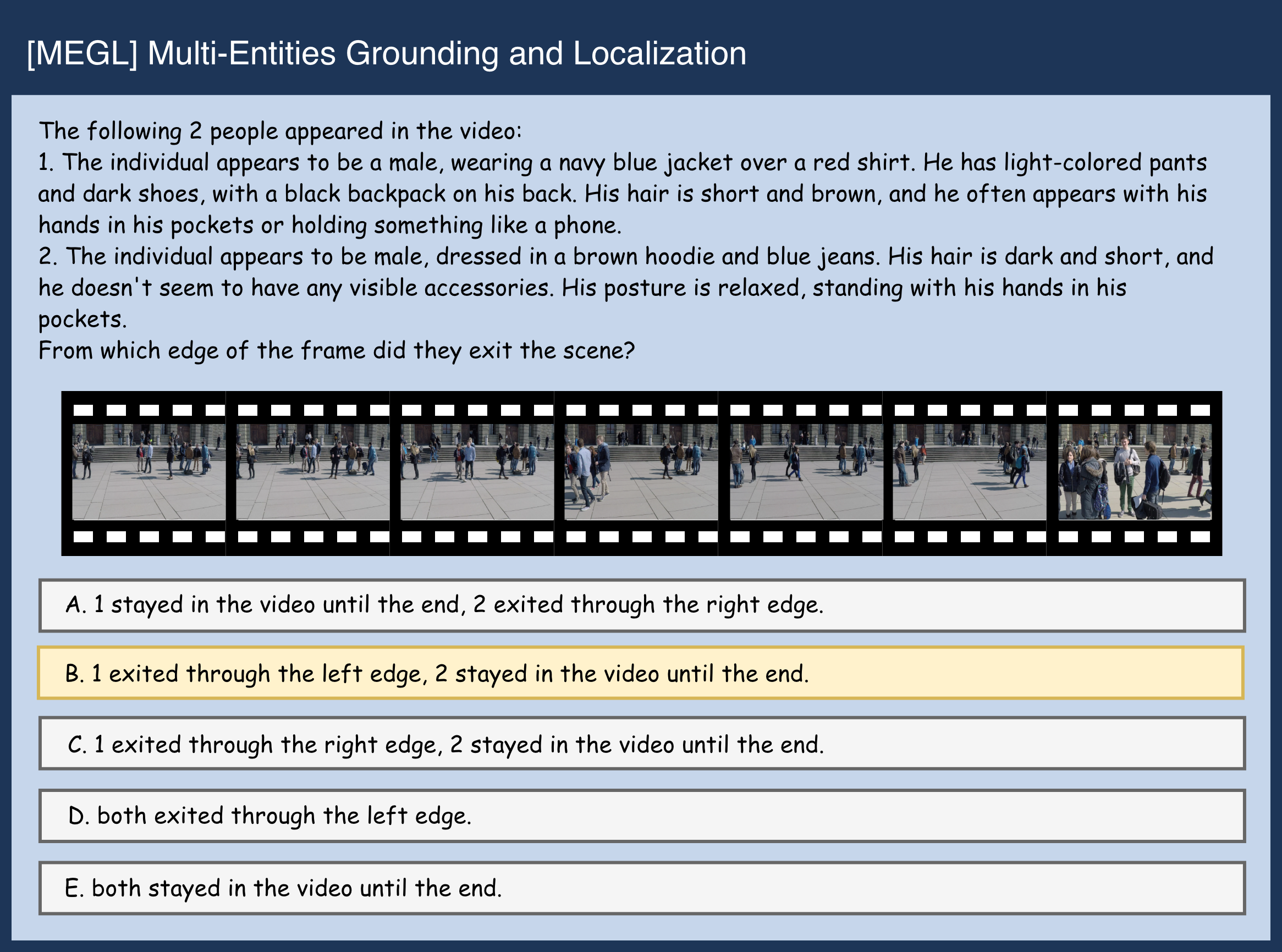}
    \caption{}
    \label{fig:megl_example} 
\end{figure*}

\begin{figure*}[!t]
    \centering
    \includegraphics[width=\textwidth]{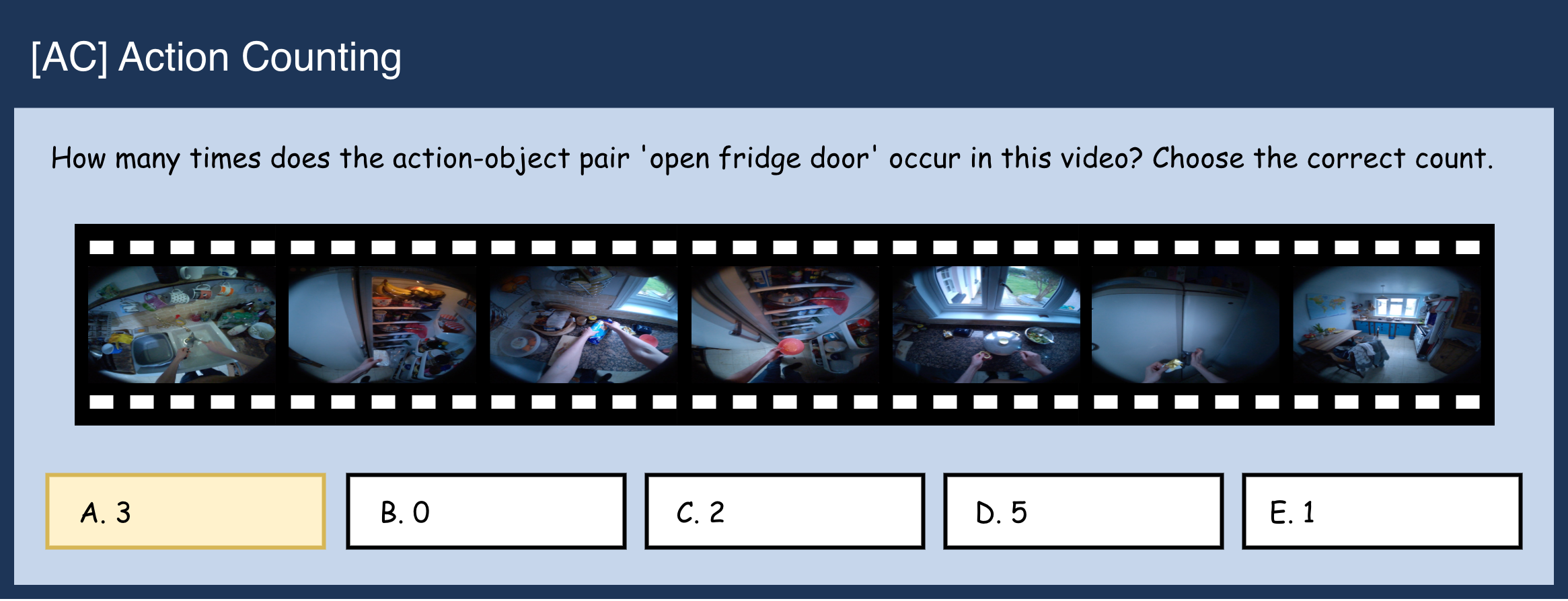}
    \caption{}
    \label{fig:ac_example} 
\end{figure*}

\begin{figure*}[t!]
    \centering
    \includegraphics[width=\textwidth]{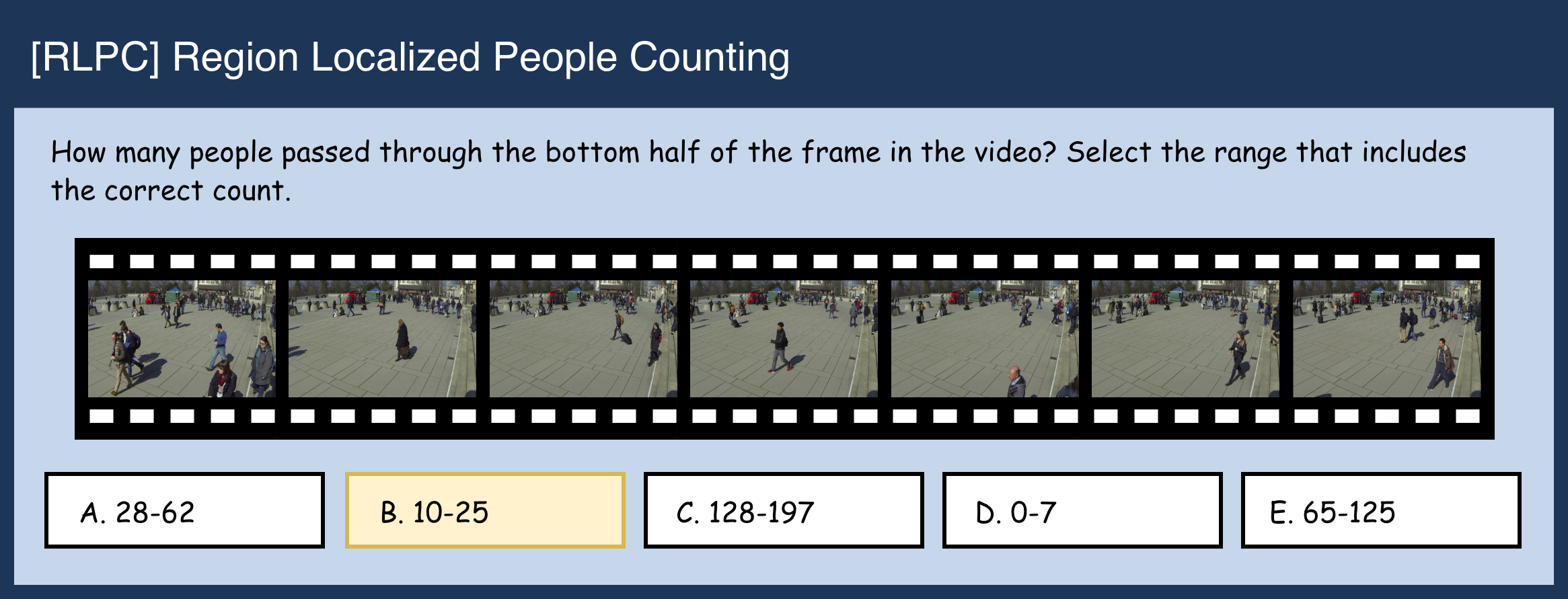}
    \caption{}
    \label{fig:rlpc_example} 
\end{figure*}

\paragraph{Multi-Entity Aggregation \& Numeracy.}
Figure~\ref{fig:megl_example} provides an example of the \textit{Multi-Entities Grounding \& Localization ([MEGL])} task, where models must verify which appearance-described individuals actually appear in the video. 
Figure~\ref{fig:ac_example} illustrates the \textit{Action Counting ([AC])} task, requiring enumeration of all instances of a specified action–object pair across the entire video. 
Finally, Figure~\ref{fig:rlpc_example} shows the \textit{Region-Localized People Counting ([RLPC])} task, where the model must count unique individuals entering through specific spatial regions.

\end{document}